\documentclass[conference]{IEEEtran}
\usepackage[utf8]{inputenc}
\usepackage[svgnames]{xcolor}
\usepackage{colortbl}
\usepackage[dvipsnames,table]{xcolor}
\usepackage{cite}
\usepackage{float}
\usepackage{amsmath,amssymb,amsfonts}
\usepackage{algorithmic}
\usepackage{graphicx}
\usepackage{subcaption}
\usepackage{textcomp}
\usepackage{bm}
\usepackage{booktabs} 
\usepackage{array}
\usepackage{makecell}
\usepackage{multirow}
\usepackage[T2A]{fontenc}
\usepackage[russian, english]{babel}
\usepackage{tempora} 
\def\BibTeX{\rm B\kern-.05em{\sc i\kern-.025em b}\kern-.08em
    T\kern-.1667em\lower.7ex\hbox{E}\kern-.125emX}

\definecolor{secgray}{gray}{0.93}
\definecolor{rowblue}{RGB}{219,234,254}

\begin{document}



\title{Perceptual Color Difference Modeling Using Machine Learning and Human Similarity Judgments}

\author{
\IEEEauthorblockN{
Elnara Kadyrgali\IEEEauthorrefmark{1},
Muragul Muratbekova\IEEEauthorrefmark{1},
Adilet Yerkin\IEEEauthorrefmark{1},
Nuray Toganas\IEEEauthorrefmark{1},
Ayan Igali\IEEEauthorrefmark{1},\\
Malika Ziyada\IEEEauthorrefmark{1},
Aruzhan Burambekova\IEEEauthorrefmark{1},
Jamaladdin Hasanov\IEEEauthorrefmark{2},
Pakizar Shamoi\IEEEauthorrefmark{1}
}

\IEEEauthorblockA{\IEEEauthorrefmark{1}
School of Information Technology and Engineering (SITE),
Kazakh-British Technical University,
Almaty, Kazakhstan
}

\IEEEauthorblockA{\IEEEauthorrefmark{2}
School of Information Technology and Engineering (SITE),
ADA University,
Baku, Azerbaijan
}

\IEEEauthorblockA{
Email: p.shamoi@kbtu.kz
}
}


\maketitle




\begin{abstract}
 Accurate assessment of color differences is essential for applications ranging from digital design to quality control. While existing color difference metrics, such as CIEDE2000, aim to approximate human perception, they may still exhibit inconsistencies with perceptual judgments. In this study, we investigate a data-driven approach to color-difference estimation based directly on human evaluations. We collect similarity judgments for 2,000 systematically generated color pairs, each rated by seven observers using a four-point ordinal scale. These judgments are then used to train regression models using different color representations, including RGB channel differences, HSI differences, and COLIBRI fuzzy linguistic categories. Experiments with five regression algorithms show that the choice of color model has a greater influence on prediction performance than the choice of regression algorithm. Using COLIBRI features alone, linear regression achieves an $R^2$ of 0.595, outperforming RGB and HSI representations, which achieve $R^2$ values of 0.479 and 0.493, respectively. The best performance is obtained by LightGBM using the combined representation, reaching an $R^2$ of 0.703. 
 The results indicate that human perceptual color differences are better captured when numerical color coordinates are complemented by graded perceptual categories, highlighting the potential of data-driven models for perceptually aligned color-difference estimation.
 
\end{abstract}



\section{Introduction}
\label{sec:intro}


Color is a fundamental visual attribute that plays a key role in describing and distinguishing image content \cite{electronics10040470, Cieplinski2001MPEG-7}. It is central to image segmentation, retrieval, and classification tasks \cite{Garcia2018Segmentation}.

Color difference estimation plays a crucial role in intelligent image processing systems \cite{he2024multiscaleslicedwassersteindistances, inproceedings}. Modern computer vision, quality control automation, image retrieval, and visualization rely on quantitative color-distance metrics. However, traditional mathematical models, such as Euclidean distance in RGB space or perceptually refined models like CIE76, CIE94, CMC $(l:c)$, and CIEDE2000, are handcrafted mathematical models designed to approximate perceptual uniformity rather than learned representations derived from human judgments. They do not always capture subtle variations as humans perceive them. In addition, CIEDE2000 is computationally heavy for large-scale CV tasks \cite{Pereira2020Efficient}.

Although CIEDE2000 remains the most widely accepted perceptual metric, it is inherently static and parametric. It does not adapt to contextual factors, presentation conditions, or observer variability. This raises a critical limitation: human color perception is nonlinear, context-dependent, and subject to cognitive and spatial influences, while most existing color-difference metrics are deterministic and globally fixed.



To investigate this problem, we conducted two controlled perceptual experiments to construct a structured dataset of human similarity judgments. A total of 2000 unique color pairs were systematically generated in the RGB space using controlled Euclidean shift levels to uniformly cover a broad spectrum of color differences. Participants evaluated perceived similarity using a four-point ordinal scale (1 -- Not similar, 2 -- Somewhat similar, 3 -- Similar, 4 -- Very similar).

The experiments were conducted under two distinct spatial configurations: (i) separated color patches and (ii) directly adjacent patches with no visible gap. This dual-condition design allows analysis of how presentation structure influences similarity perception.

The collected perceptual dataset enables:
\begin{itemize}
    \item Learning perceptual similarity thresholds via logistic regression.
    \item Quantifying inter-observer agreement using Fleiss' Kappa.
    \item Comparing human similarity judgments with analytical metrics such as CIEDE2000.
\end{itemize}

Unlike traditional studies that validate fixed color-difference formulas, this work frames perceptual color difference as a supervised similarity-learning problem. By integrating human-labeled similarity data with computational intelligence techniques, we move toward adaptive, context-aware perceptual metrics that bridge the gap between mathematical color spaces and human visual cognition.

The main contributions of this study are as follows:
\begin{itemize}
    \item Construction of a structured perceptual similarity dataset based on 2000 systematically generated RGB color pairs.
    \item Provision of regression coefficients that enable perceptual color difference to be calculated directly from two RGB color values and provide interpretability by quantifying the contribution of individual color features.
    \item Comparative analysis between human similarity judgments and conventional color-difference formulas.
    \item Empirical demonstration that spatial presentation conditions significantly affect perceptual similarity thresholds.
\end{itemize}


The structure of the paper is as follows. Section I is this introduction. Section II reviews related work on perceptual color-difference metrics.  Section III presents the methodology.
Section IV describes the experimental design and data generation process.  Section V reports experimental results and analysis. 
Section VIII concludes the paper and provides ideas for future work.

\section{Related Work}


\begin{table*}[!t]
\centering
\renewcommand{\arraystretch}{1.25}
\caption{Summary of Prior Studies on the Effect of Sample Separation on color-Difference Evaluation. Studies are grouped by stimulus type. The number of observers ranged from 4 to 46 across studies; see original sources for details. NS\,=\,no separation; —\,=\,not reported in original source.}
\label{tab:color_difference_separation}
\scriptsize
\setlength{\tabcolsep}{4pt}
\begin{tabular}{>{\raggedright}p{2.5cm} >{\raggedright}p{2.6cm} >{\centering}p{1.1cm} >{\centering}p{1.6cm} >{\raggedright}p{2.2cm} >{\raggedright\arraybackslash}p{6cm}}
\toprule
\textbf{Study} & \textbf{Separation conditions} & \textbf{Color pairs} & \textbf{Color difference range} & \textbf{Assessment method} & \textbf{Key finding (separation effect)} \\ \midrule
\rowcolor{secgray}\multicolumn{6}{l}{\textit{Vision science foundations}} \\
Boynton et al., 1977~\cite{Boynton1977GapEffect}& Juxtaposed; with gap & — & Threshold & Threshold detection & Foundational gap effect study: gap impaired luminance discrimination but improved chromatic discrimination; explained by contour enhancement and spatial averaging \\ \hline
Sharpe \& Wyszecki, 1976~\cite{Sharpe1976ProximityFI} & Juxtaposed; with gap & — & Threshold \& supra & Ratio comparison; liminal det.; color matching & Separation impairs lightness more than chromaticness; proposed a proximity factor for color-difference formulas \\ \hline
Eskew, 1989~\cite{Eskew1989TheGE} & Juxtaposed; narrow gap (isoluminant fill) & — & Threshold & Threshold detection & Luminance or chromatic gap prevents spatial integration, enhancing chromatic sensitivity; gap had little effect for flashed stimuli \\ \hline
Danilova \& Mollon, 2006~\cite{Danilova2006SeparatedColours} & Juxtaposed; separated up to 10° & — & Threshold & 2AFC threshold & Discrimination optimal at small gap; thresholds rose moderately with increasing separation; gap effect more pronounced in parafovea \\ \hline
\rowcolor{secgray}\multicolumn{6}{l}{\textit{Psychophysical experiments --- surface \& printed colors}} \\
Witt, 1990~\cite{Witt1990ParametricEO} & Gap vs.\ no gap & — & Threshold & Threshold detection & Gap raised the color-difference threshold consistently, especially for darker CIE color centres \\ \hline
Guan \& Luo, 1999~\cite{guan_luo_1999} & Hairline \& 3-inch gap & 75 & $\approx$3\,$\Delta E^*$ (CIELAB) & Gray-scale; pair comparison & Hairline separation gave 8\% larger perceived $\Delta E$ than 3-inch gap; both psychophysical methods yielded similar results \\ \hline
Ting Xu et al., 2019~\cite{Xu2019EffectOP} & With gap; without gap & 460 & 1, 2, 4, 8\,$\Delta E^*$ (CIELAB) & Gray-scale & Gap effect more pronounced at smaller $\Delta E$; CAM16-UCS outperformed CIEDE2000, CIE94, CMC, CIELAB \\ \hline
Mirjalili et al., 2019~\cite{Fereshteh_CDMirjalili} & No separation (hairline as reference) & 1012 & 1, 2, 4, 8\,$\Delta E^*$ (CIELAB) & Gray-scale & Clear separation effect vs.\ hairline reference; proposed $\Delta E_{\text{NS}}$ (modified CIEDE2000); for $\Delta E_{00}{<}9.1$ larger magnitude increases perceived lightness border \\ \hline
Brusola et al., 2019~\cite{Brusola2019} & NS; 0.5\,mm black gap; 3\,mm white gap & — & Threshold & Strip-pair comparison & Separation type had little effect on chromaticity-discrimination ellipse shape; strip method correlated well with pair comparison \\ \hline
\rowcolor{secgray}\multicolumn{6}{l}{\textit{Psychophysical experiments --- display colors (CRT\,/\,LCD)}} \\
Cui et al., 2001~\cite{Cui2001ColourDifferenceEU} & NS; 1-px; 2-px; large gap & — & — & Gray-scale & Gap size had minor effect on overall $\Delta E$ but shifted lightness/chromatic weighting; clear difference between NS and any gap condition \\ \hline
Xu \& Yaguchi, 2005~\cite{Xu_Haisong_Hirohisa} & Hairline separation & — & Threshold to large supra & Interleaved staircase; constant stimuli & Inter-observer variability ${\approx}$40\% higher at small $\Delta E$; CIEDE2000 outperformed CMC, CIE94, CIELAB across full threshold-to-suprathreshold range \\ \hline
Qiang Xu et al., dep2022~\cite{Xu2022APC} & .Separation; no separation & 1120 & 4, 8\,$\Delta E^*$ (CIELAB) & Gray-scale & CIEDE2000 best for gap pairs; all models worse for NS pairs; CMF had negligible influence on $\Delta E$ values \\ \hline
\rowcolor{rowblue}
\textbf{This study} & \textbf{Gap; no separation} & \textbf{2000} & \textbf{0–222 (Euclidean RGB)\,$\Delta E^*$} & \textbf{4-point ordinal similarity scale (web-based)} & \textbf{First to examine separation effect across wide $\Delta E$ range 0–222 (Euclidean RGB) on display; develops predictive model for both NS and gap conditions} \\
\bottomrule
\end{tabular}
\end{table*}

Color perception has long been studied as an important aspect of human cognition and visual understanding. Early studies by Berlin and Kay demonstrated that people across different cultures tend to identify similar focal colors, whereas category boundaries vary considerably \cite{berlin1969basic}. These findings suggest that color perception is shaped not only by physical stimuli but also by perceptual and cognitive mechanisms. Building on this idea, color difference research focuses on understanding how humans perceive similarities and differences between colors.

Xu et al. \cite{Xu_Haisong_Hirohisa} investigated color-difference perception across threshold and suprathreshold levels and demonstrated that perceptual discrimination is nonlinear. Their results showed that CIEDE2000 provides better agreement with human judgments than alternative color-difference models, contributing to the development of modern perceptual color metrics. Furthermore, studies on dental shade matching \cite{Dental_Oscar} highlighted that, although instrumental measurements provide greater objectivity, subjective human perception remains an essential component of color evaluation. Similarly, Huang et al. \cite{huang2025acceptable} demonstrated that observer experience and lighting conditions significantly influence color-difference perception and improve the prediction accuracy of perceptual color-difference models.

The importance of color difference extends beyond perception itself. Wang et al. \cite{Wang2019} showed that increasing color diversity among targets and reducing target--distractor distinctiveness both impair tracking performance in multiple object tracking tasks. Their results indicate that visual performance depends primarily on perceptual distinctiveness rather than on the number of colors alone.

To quantify perceptual color differences, a variety of color spaces and distance metrics have been developed. Rodriguez et al. \cite{rodriguez2018assessment} demonstrated that CIEDE2000 outperforms alternative measures such as SAM and RMSE in image quality evaluation. Despite its relatively high computational complexity \cite{aligholi2015automated}, CIEDE2000 remains widely used because of its strong correspondence with human visual perception \cite{melgosa2017color}. In addition, as an alternative to fixed formula-based color-difference evaluation, an ANFIS was proposed to model color closeness \cite{Hasanov2021}.

Although earlier color-difference formulas improved prediction accuracy, they remained largely dependent on the CIELAB space and were limited in modeling viewing conditions and perceptual adaptation. This motivated the development of perceptually uniform color appearance models, evolving from CIECAM02 and CAM02-UCS to CAM16 and CAM16-UCS \cite{luo2006uniform, li2017comprehensive}. Similarly, Mirjalili et al. \cite{mirjalili2019color} showed that viewing conditions significantly influence perceived color differences, while Konovalenko et al. \cite{konovalenko2021prolab} proposed the proLab color space to improve perceptual uniformity while preserving linear color relationships.

The geometric properties of color spaces have also received considerable attention. Misue \cite{misue2020} developed an interactive visualization tool for exploring CIE76, CIE94, and CIEDE2000, demonstrating substantial differences in perceptual distance regions, particularly for highly saturated colors. More broadly, Lissner and Urban \cite{lissner2012toward} argued that no single color space can effectively support all perception-based image processing tasks and emphasized the need for unified frameworks capable of handling both threshold and suprathreshold color differences.

Color-difference information is widely used in image processing and computer vision tasks. Niu et al. \cite{Niu2018} proposed the Color Contrast Similarity and Color Value Difference (CSVD) metric for evaluating color correction quality and demonstrated stronger agreement with subjective assessments than numerous existing image quality metrics.

Several studies have incorporated perceptually uniform color spaces into image quality assessment and saliency detection. Achanta et al. \cite{achanta2009frequency}, Lee et al. \cite{lee2015towards}, and Zhang et al. \cite{zhang2013sdsp} employed CIELAB-based representations to better reflect human visual perception. Lee et al. \cite{lee2014towards, lee2015towards} further demonstrated that incorporating lightness, hue, and chroma improves color image quality assessment. In addition, Shi et al. \cite{shi2024no} proposed a no-reference sharpness assessment model based on color-difference variations, achieving improved prediction accuracy with lower computational complexity. Similarly, Liu et al. \cite{Liu2019} showed that perceptually motivated color differences in the Lab* space contribute to more accurate salient object detection.

Beyond image analysis, color-difference models have been applied in visualization, design, and optimization tasks. Szafir et al. \cite{Szafir_Danielle} demonstrated that color discriminability depends not only on color differences but also on visualization context, leading to probabilistic models that improve color encoding in data visualization.

Misue \cite{misue2016design} developed a CIELAB-based tool for constructing perceptually consistent sequential, divergent, and qualitative color schemes, highlighting challenges associated with gamut constraints and perceptual distinguishability. Similarly, Kosesoy et al. \cite{kosesoy2025novel} employed CIEDE2000 in a nature-inspired color palette generation framework, demonstrating its usefulness for design-oriented applications. Wu et al. \cite{wu2021green} further showed that color-difference metrics can support environmentally sustainable printing by reducing ink consumption while preserving perceptual print quality.

Overall, previous studies have significantly advanced the understanding of color difference perception, perceptually uniform color spaces, and their applications in image processing, visualization, and design \cite{Xu_Haisong_Hirohisa, li2017comprehensive, konovalenko2021prolab, Niu2018, Szafir_Danielle}. Existing research has demonstrated that color differences influence human perception, attention, and decision-making, while modern color-difference models have improved the prediction of perceptual similarity under a variety of viewing conditions \cite{Wang2019, rodriguez2018assessment, mirjalili2019color, huang2025acceptable}. However, most current approaches focus primarily on perceptual distinguishability, visual quality assessment, or color appearance consistency, with relatively limited attention given to the emotional meaning associated with color differences. Furthermore, although observer-dependent factors have been acknowledged in color perception studies \cite{Dental_Oscar, huang2025acceptable}, the subjective and uncertain nature of color--emotion associations remains insufficiently represented in existing color-difference frameworks. This limitation becomes particularly important in multimodal environments, where visual and textual information jointly contribute to emotional interpretation. Therefore, there is a need for approaches capable of modeling color--emotion relationships while explicitly accounting for uncertainty and subjectivity through fuzzy and multimodal representations.

\section{Methodology}

In this study, we conduct a perceptual experiment to understand how people judge color differences and how spatial separation influences their perception. The experiment focuses on collecting subjective similarity assessments, which are then used to build a training dataset for a machine learning-based model and to compare human judgments with existing color difference measures. We also evaluate different feature representations and learning algorithms to determine which combination best reflects human perception.

\begin{figure*}[ht!]
    \centering
    \begin{subfigure}[t]{0.48\textwidth}
        \centering
        \includegraphics[width=\linewidth]{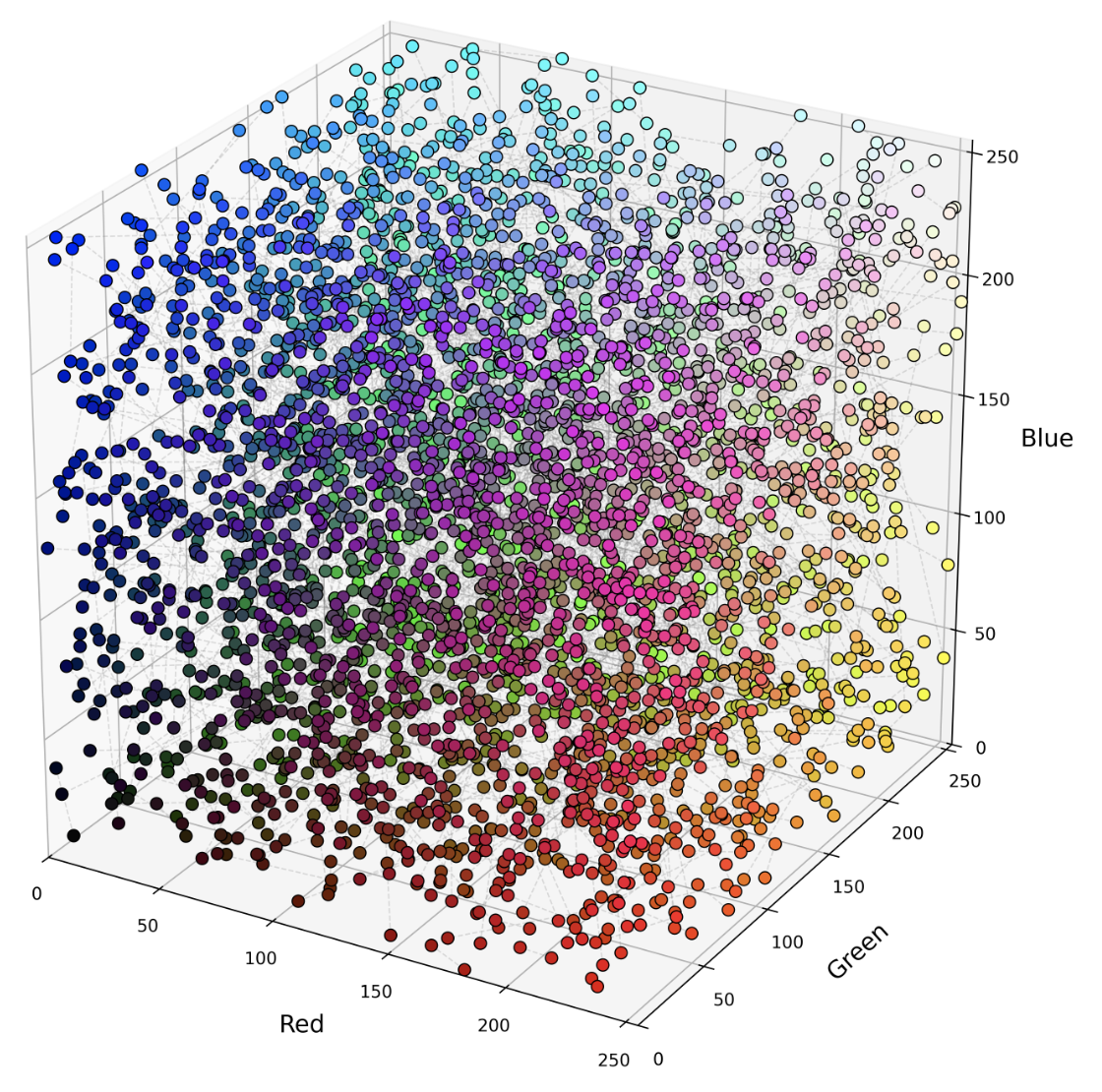}
        \caption{Color pairs in RGB representation}
        \label{fig:colorcube}
    \end{subfigure}
    \hfill
    \begin{subfigure}[t]{0.48\textwidth}
        \centering
        \includegraphics[width=\linewidth]{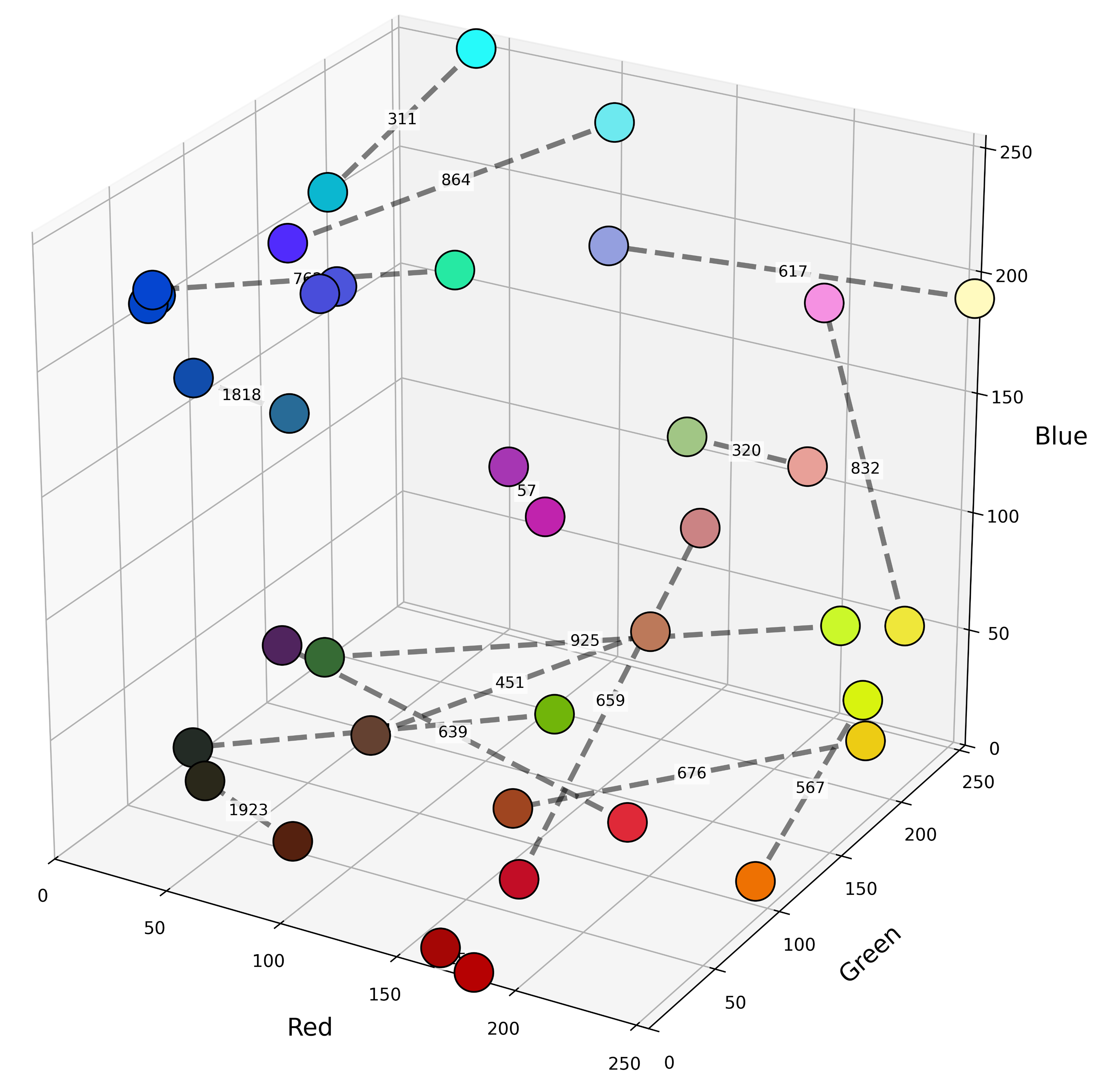}
        \caption{Color pairs example}
        \label{fig:colorcube_pairs}
    \end{subfigure}
    \caption{Color pairs visualization}
    \label{fig:rgbcube}
\end{figure*}

\subsection{Data collection}

The RGB color model constitutes the de facto standard for color representation in contemporary digital display systems. Accordingly, all stimuli were specified in the RGB format to ensure reproducibility and facilitate comparability across experimental conditions and setups. The RGB color space encompasses 16,777,216 distinct colors, which yields more than 1.4 × 10\^14 possible unordered color pairs. A comprehensive, exhaustive evaluation of such a large combinatorial space is both computationally intractable and experimentally unmanageable. Therefore, it is necessary to employ an efficient sampling strategy that selects a representative subset of color pairs while preserving the statistical robustness and reliability of the experimental outcomes.


For the experiment, we generated 2000 distinct color pairs (see Fig. \ref{fig:colorcube}) constructed to systematically span a broad range of perceptual color differences. Each color was specified in the RGB color space, with channel values sampled from the discrete interval [0, 255].

Initially, a random reference color $\mathbf{C}_1(R_1, G_1, B_1)$ was sampled from the RGB color space. Subsequently, a second color $\mathbf{C}_2(R_2, G_2, B_2)$ was iteratively selected such that the Euclidean distance between $\mathbf{C}_1$ and $\mathbf{C}_2$ in RGB space matched a predefined shift magnitude within a specified tolerance. The color difference $\mathbf{D}(\mathbf{C}_1,\mathbf{C}_2)$ between two colors $\mathbf{C}_1$ and $\mathbf{C}_2$ was calculated as:

\begin{equation}
D(\mathbf{C}_1, \mathbf{C}_2) = \sqrt{(R_1 - R_2)^2 + (G_1 - G_2)^2 + (B_1 - B_2)^2}.
\end{equation}

A uniform sampling of color differences across the designated range was achieved by varying the shift levels from 0 to 220 and generating independent color pairs for each shift level (some examples illustrated in Fig. \ref{fig:colorcube_pairs}). The resulting color distances ranged roughly from 1.00 to 221.79 because of the applied tolerance and the discrete nature of the RGB space. Color differences in the range greater than 220 were excluded, as extremely large differences produce pairs that are trivially unlike each other and therefore do not enable meaningful similarity-based categorization in the experiment.

Due to the wide and finely spaced range of color difference magnitudes achieved with this controlled sampling approach, the experiment was able to assess perceptual judgments for both very small and very large color differences. To support reproducibility and additional analysis, all generated color pairs were saved along with their RGB values, hexadecimal codes, and precise computed distances.

\label{sec:methodology}


\subsection{Experiment Design}
\label{sec:exp_settings}

\begin{figure*}[t!]
    \centering
    \begin{subfigure}[t]{0.48\textwidth}
        \centering
        \includegraphics[width=\linewidth]{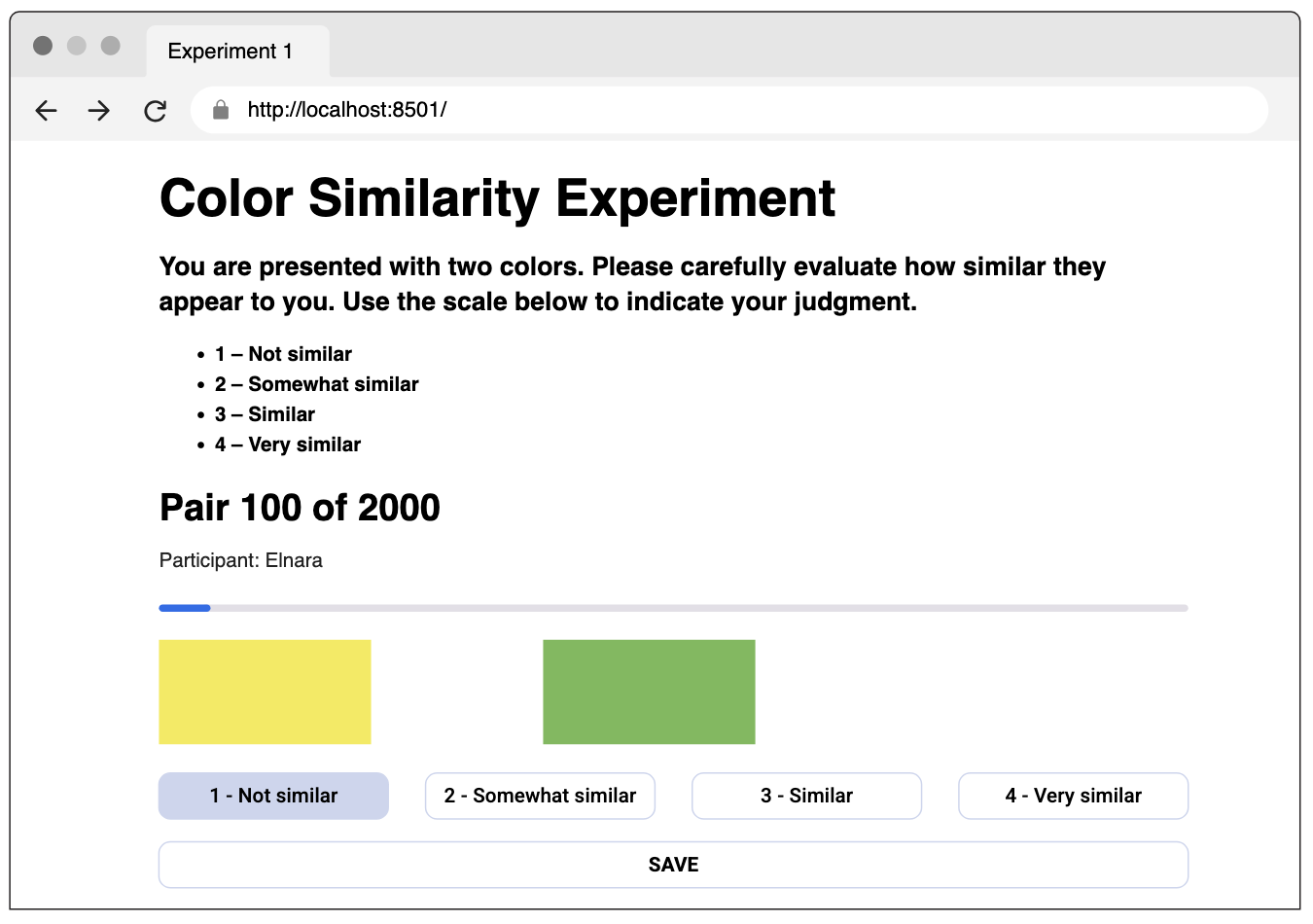}
        \caption{Experiment 1: Color pairs with spatial separation}
        \label{fig:experiment_a}
    \end{subfigure}
    \hfill
    \begin{subfigure}[t]{0.48\textwidth}
        \centering
        \includegraphics[width=\linewidth]{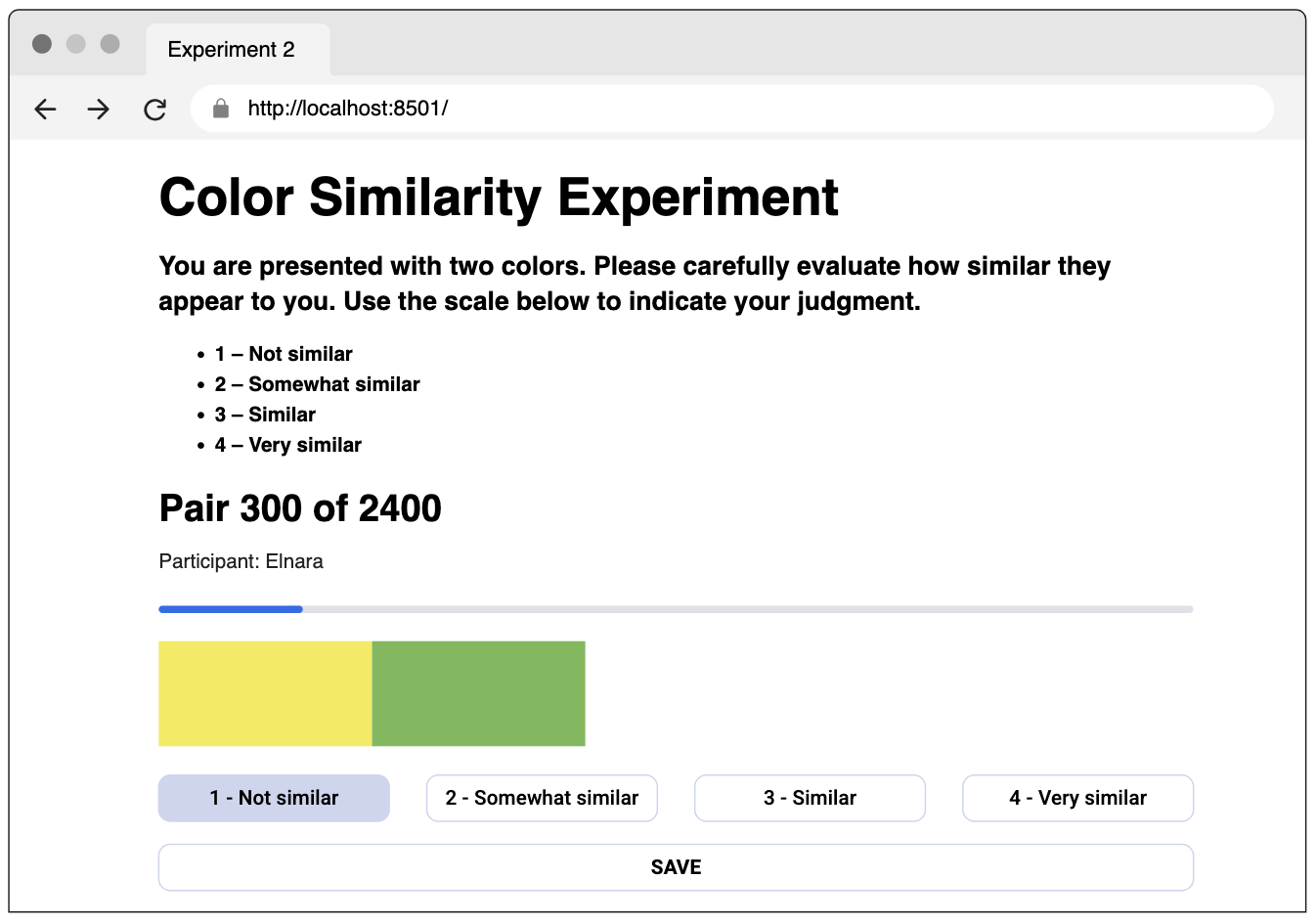}
        \caption{Experiment 2: Color pairs with no separation}
        \label{fig:experiment_b}
    \end{subfigure}
    \caption{Interfaces used in the color similarity judgement experiments}
    \label{fig:experiment}
\end{figure*}

The study includes two controlled experiments on color similarity judgements to examine how people subjectively perceive color difference/similarity under two different presentation conditions, as shown in Fig. \ref{fig:experiment}.

A web-based interface was used to manage both experiments, guaranteeing automated data collection and uniform stimulus presentation. Each experiment started with instructions, after which participants had to enter their names in order to proceed. Prior to the main experimental phase, a practice trial was provided to familiarize participants with the task and the rating scale (responses from trial were not included in the analysis).

Participants rated the perceived similarity of color pairs using a pre-described  four-point ordinal scale:
\begin{enumerate}
    \item Not similar
    \item Somewhat similar
    \item Similar
    \item Very similar.
\end{enumerate}

During the first experimental phase (see Fig. \ref{fig:experiment_a}), participants were presented with pairs of colors drawn from a prepared stimulus set and displayed simultaneously as two spatially separated color patches of equal size. The order of color pairs was randomized independently for each participant. The system recorded the participant identifier, the color pair identifier, the predefined color difference (Euclidian distance) and the selected similarity rating.

The same stimulus pool and rating scale were used in the second experiment, to maintain comparability with Experiment 1. To assess intra-participant consistency, a subset of 400 color pairs was randomly selected from the original stimulus set and duplicated. These duplicated pairs were merged with the full stimulus set, and the resulting list was randomly shuffled so that repeated presentations occurred at unpredictable positions within the sequence. In contrast to Experiment 1, the two colors in each pair were displayed as adjacent color patches with no spatial separation (see Fig. \ref{fig:experiment_b}), to facilitate direct perceptual comparison and to reduce potential visual segmentation effects. The other experimental settings of the second experiment were identical to Experiment 1. Right after the completion of each experiment, all responses were automatically saved for subsequent statistical analysis.

Together, the two experiments made it possible to analyse response consistency through repeated stimulus evaluations and to conduct a systematic investigation of color similarity perception under various presentation conditions. 


\subsection{Spatial Separation Conditions}
Experimental studies on color-difference perception differ not only in the employed 
color-difference models but also in the spatial arrangement of the color samples being 
compared. The spatial relationship between compared 
color samples plays a critical role in visual perception. Three principal presentation 
conditions are commonly distinguished: hairline separation, separation with a visible 
gap, and no-separation.

Hairline separation refers to a configuration in which two color samples are placed 
side by side with only a minimal boundary between them - sufficient to perceptually 
segment the samples. This condition has 
been widely adopted as a standard reference in classical color-difference studies and 
in the development of color-difference formulae, as it limits spatial interaction 
effects while maintaining clear sample boundaries~\cite{guan_luo_1999, 
Cui2001ColourDifferenceEU}.

Separation with a visible gap involves a clearly noticeable spatial distance between 
color samples, typically filled with a neutral background. This configuration reduces 
spatial interaction between adjacent colors and is used to investigate the 
influence of separation distance on perceived color differences. Under such conditions, 
perceptual evaluations have been reported to show greater consistency with conventional 
color-difference models compared to the no-separation condition~\cite{guan_luo_1999, 
Cui2001ColourDifferenceEU}.

In contrast, the no-separation condition is characterized by the direct adjacency of color samples with no boundary between them. This configuration is 
representative of many real-world scenarios, such as printed images and digital 
displays. Experimental studies have demonstrated that no-separation conditions 
systematically alter the perceived balance between lightness and chromatic differences: 
in some conditions, observers show increased sensitivity to lightness variations, 
leading to deviations from the predictions of traditional color-difference 
formulae~\cite{Cui2001ColourDifferenceEU, Xu2019EffectOP, Xu2022APC}.

These three conditions differ not only in their physical configuration but also in how they affect the relative weighting of lightness and chromatic components in perceived 
color difference. The manner in which color pairs are presented has been shown to 
systematically influence color-difference judgments across a range of materials and 
viewing conditions~\cite{Sharpe1976ProximityFI, Eskew1989TheGE, Fereshteh_CDMirjalili}.

Table~\ref{tab:color_difference_separation} summarizes the experimental 
characteristics of previous studies and highlights the gap addressed 
by the present work: to our knowledge, no prior study has examined the separation effect 
across a wide $\Delta E$ range (0--220) on a display medium. 

\subsection{Color Models}


Three complementary color models are considered in this study: RGB, HSI, and COLIBRI. 

RGB represents color using three numerical channels (Red, Green, Blue) and is the standard representation for digital images, whereas HSI separates color information into perceptually meaningful hue, saturation, and intensity components \cite{Muratbekova2026}. In contrast, COLIBRI provides a fuzzy linguistic representation of color, describing hue, saturation, and intensity through perceptually motivated categories with gradual membership rather than strict numerical boundaries \cite{COLIBRI2025}.




\subsection{Machine Learning algorithms}
We implement five machine learning (ML) algorithms for the regression task of computing the color difference between two colors: \textit{Linear Regression, Decision Tree, Random Forest, LightGBM, and XGBoost}. These models were selected to cover different levels of model complexity, ranging from a simple linear baseline to nonlinear tree-based and ensemble methods. This allows us to examine whether perceptual color differences can be represented by a relatively simple relationship or require more complex nonlinear modeling.

\subsubsection{Linear Regression}
Linear Regression is a baseline ML algorithm \cite{hastie2009elements} that models the relationship between features and the target using interpretable coefficients, which aligns with our approach to perceptual human color difference. The prediction of the model can be expressed as
\begin{equation}
    \hat{y} = \beta_0 + \sum_{j=1}^{p} \beta_j x_j ,
\end{equation}
where $x_j$, $j = 1,\dots,p$, are the input features, $\beta_0$ is the intercept, $\beta_j$ are the learned coefficients, and $\hat{y}$ is the predicted value.

\subsubsection{Decision Tree}
A decision tree \cite{breiman1984cart} is a supervised ML algorithm that adopts a hierarchical tree structure. At each node, the algorithm selects a feature and a threshold that provide the best split of the data. The optimal split is obtained by minimizing the weighted impurity of the resulting subsets,
\begin{equation}
    s^{*} = \arg\min_{s}
    \left[
        \frac{n_{\mathrm{left}}}{n}\, L(D_{\mathrm{left}})
        + \frac{n_{\mathrm{right}}}{n}\, L(D_{\mathrm{right}})
    \right],
\end{equation}
where $s$ denotes a candidate split, $D_{\mathrm{left}}$ and $D_{\mathrm{right}}$ are the resulting subsets containing $n_{\mathrm{left}}$ and $n_{\mathrm{right}}$ samples, $n = n_{\mathrm{left}} + n_{\mathrm{right}}$, and $L$ is the impurity function, taken as the within-node mean squared error for regression.

\subsubsection{Random Forest}
Random Forest \cite{breiman2001random} is an ensemble ML algorithm that builds multiple decision trees on bootstrap samples with randomized feature selection, and whose overall prediction is obtained by averaging the predictions of the individual trees. The final prediction for a regression task is given by
\begin{equation}
    \hat{y} = \frac{1}{B}\sum_{b=1}^{B} f_b(x),
\end{equation}
where $B$ is the number of trees and $f_b(x)$ is the prediction of the $b$-th tree.

\subsubsection{LightGBM}
LightGBM (Light Gradient Boosting Machine) \cite{ke2017lightgbm} is a gradient boosting decision tree algorithm designed for computational efficiency and scalability. While traditional gradient boosting methods grow trees level-wise, LightGBM uses a leaf-wise strategy, splitting the leaf that provides the largest reduction in the objective function. As a boosting algorithm, the model is updated iteratively according to
\begin{equation}
    F_m(x) = F_{m-1}(x) + \eta\, f_m(x), \qquad m = 1,\dots,M,
\end{equation}
where $F_{m-1}(x)$ is the current model, $f_m(x)$ is the newly added tree, $\eta$ is the learning rate, and $M$ is the number of boosting iterations.

\subsubsection{XGBoost}
XGBoost (eXtreme Gradient Boosting) \cite{chen2016xgboost} is an ML algorithm that builds decision trees sequentially, where each new tree aims to correct the errors made by the previous ones. In contrast to regular gradient boosting, XGBoost incorporates explicit regularization to reduce overfitting, uses second-order gradient information for optimization, and efficiently handles sparse and missing data. At iteration $t$, the training objective is formulated as
\begin{equation}
    \mathcal{L}^{(t)}
    =
    \sum_{i=1}^{n}
    l\!\left(y_i,\; \hat{y}_i^{(t-1)} + f_t(x_i)\right)
    + \Omega(f_t),
\end{equation}
\begin{equation}
    \Omega(f_t) = \gamma T + \frac{1}{2}\lambda \lVert w \rVert^{2},
\end{equation}
where $l$ is the prediction loss, $\hat{y}_i^{(t-1)}$ is the prediction of the model after $t-1$ iterations, and $\Omega(f_t)$ is the regularization term penalizing model complexity through the number of leaves $T$ and the leaf weights $w$, with $\gamma$ and $\lambda$ as regularization parameters.

\subsection{Color Difference Metrics}

To quantify the perceptual difference between color pairs, several widely used color difference formulas are used to evaluate the proposed method and compare with them. 

\subsubsection{CIE76 (CIELAB Euclidean Distance) }
The CIE76 metric defines color difference as the Euclidean distance between two points in the CIELAB color space. For two colors represented by $(L_1^*, a_1^*, b_1^*)$ and $(L_2^*, a_2^*, b_2^*)$, the color difference is given by \cite{CIE1976}:
\begin{equation}
\Delta E_{ab}^* = \sqrt{(L_2^* - L_1^*)^2 + (a_2^* - a_1^*)^2 + (b_2^* - b_1^*)^2}.
\end{equation}
Although simple and computationally efficient, this formulation assumes perceptual uniformity of the CIELAB space, which has been shown to be inaccurate in regions of high chroma and low lightness.

\subsubsection{CIE94}
To address perceptual non-uniformities of CIE76, the CIE94 model introduces separate weighting of lightness, chroma, and hue components \cite{CIE1995}:
\begin{equation}
\Delta E_{94}^* = \sqrt{
\left(\frac{\Delta L^*}{k_L S_L}\right)^2 +
\left(\frac{\Delta C_{ab}^*}{k_C S_C}\right)^2 +
\left(\frac{\Delta H_{ab}^*}{k_H S_H}\right)^2 }.
\end{equation}
Here, $\Delta L^*$, $\Delta C_{ab}^*$, and $\Delta H_{ab}^*$ denote the differences in lightness, chroma, and hue, respectively, while $S_L$, $S_C$, and $S_H$ are weighting functions dependent on the reference color. The parameters $k_L$, $k_C$, and $k_H$ account for viewing conditions.


\subsubsection{CIEDE2000}
The CIEDE2000 formula further improves perceptual uniformity by introducing nonlinear corrections and a rotation term accounting for chroma--hue interactions \cite{Sharma2005CIEDE2000}:
\begin{equation}
\begin{aligned}
\Delta E_{00}
= \Bigg[
&\left(\frac{\Delta L'}{k_L S_L}\right)^2
+\left(\frac{\Delta C'}{k_C S_C}\right)^2 \\
&+\left(\frac{\Delta H'}{k_H S_H}\right)^2 \\
&+R_T
\left(\frac{\Delta C'}{k_C S_C}\right)
\left(\frac{\Delta H'}{k_H S_H}\right)
\Bigg]^{\frac{1}{2}} .
\end{aligned}
\end{equation}
The corrected terms $\Delta L'$, $\Delta C'$, and $\Delta H'$ reflect perceptual adjustments in lightness, chroma, and hue, while $R_T$ models the interaction between chroma and hue in the blue region of the color space.

\subsubsection{CMC ($l:c$)}
The CMC model introduces adjustable parameters that control the relative weighting of lightness and chroma differences \cite{Clarke1984CMC}:
\begin{equation}
\Delta E_{CMC} = \sqrt{
\left(\frac{\Delta L^*}{l S_L}\right)^2 +
\left(\frac{\Delta C_{ab}^*}{c S_C}\right)^2 +
\left(\frac{\Delta H_{ab}^*}{S_H}\right)^2 }.
\end{equation}
The parameters $l$ and $c$ determine the tolerance for lightness and chroma variations, respectively, while $S_L$, $S_C$, and $S_H$ are weighting functions dependent on the reference color. In particular, $S_L$ and $S_C$ are functions of lightness and chroma, whereas $S_H$ additionally depends on the hue angle through a trigonometric term and a chroma-dependent factor $F$, defined as $F = \sqrt{\frac{C^{*4}}{C^{*4} + 1900}}$. The hue weighting term $T$ is computed as a piecewise function of the hue angle. This formulation allows for application-specific perceptual tuning, with common parameter settings such as $l:c = 2:1$ for acceptability and $1:1$ for perceptibility.

\subsubsection{Judd--Hunter (NBS)}
An early perceptually motivated model expresses color difference using weighted luminance and chromatic components \cite{Judd1948}:
\begin{equation}
\Delta E_{AN} = \sqrt{(\Delta L)^2 + (\Delta A)^2 + (\Delta B)^2},
\end{equation}
with
\begin{equation}
L = 9.2Y, \quad A = 40(X - Y), \quad B = 16(Y - Z).
\end{equation}
A refined NBS formulation incorporates luminance-dependent chromatic sensitivity:
\begin{equation}
\Delta E_{NBS} = f_g \sqrt{\left[221 Y^{1/4} \sqrt{(\Delta \alpha)^2 + (\Delta \beta)^2}\right]^2 + \left[k(\Delta Y^{1/2})\right]^2}.
\end{equation}

\label{sec:experiment}



\section{Results}
\label{sec:results}
\subsection{Our Model}

\begin{figure*}
    \centering
    \includegraphics[width=0.85\linewidth]{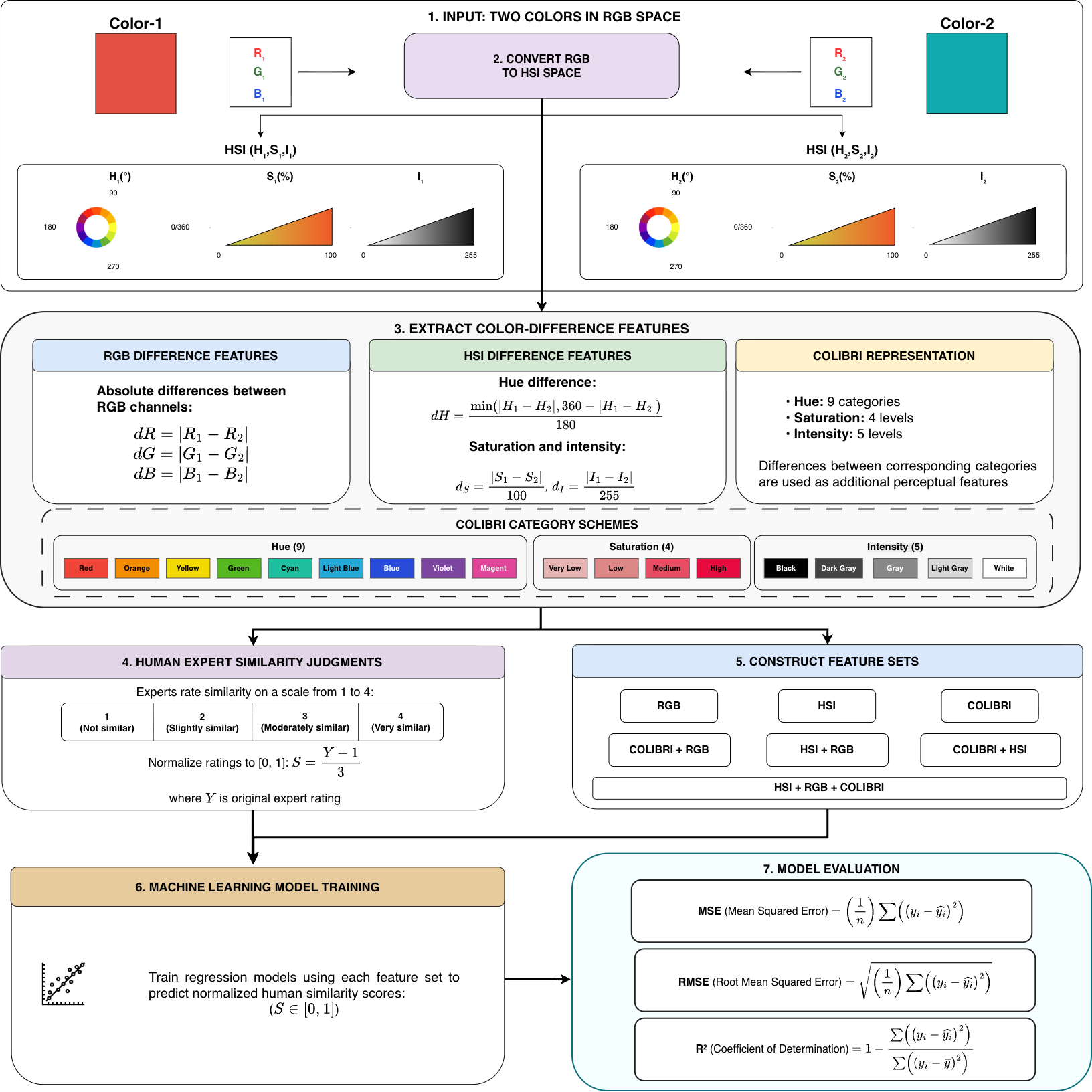}
    \caption{Proposed framework for perceptual color difference modeling }
    \label{fig:exp_methodological}
\end{figure*}

Figure~\ref{fig:exp_methodological} illustrates the overall workflow of the proposed color difference/similarity prediction framework. Starting from a pair of RGB colors, multiple feature representations are extracted and used to train machine learning models that predict human judgments.

\subsubsection{Dataset}

The training dataset was constructed from the annotations collected during the two perceptual experiments described in Section~\ref{sec:exp_settings}. Each sample consists of a pair of RGB colors and the corresponding similarity score assigned by human experts on a four-point scale , where 1 denotes ``not similar'' and 4 denotes ``very similar''. Table~\ref{tab:dataset_summary} summarizes the main characteristics of the resulting dataset.

The final dataset contains 2,000 unique color pairs evaluated by seven human experts, resulting in 28,600 individual similarity ratings. 


To formulate the task as a regression problem, the four-point ordinal similarity scale was treated as approximately equally spaced using

\begin{equation}
\label{mapp_regression}
S=\frac{Y-1}{3},
\end{equation}

where $Y$ is the original expert rating and $S$ is the normalized similarity score used as the prediction target.

Under this modeling assumption, the original ratings were linearly transformed to the interval $[0,1]$ according to Eq. \ref{mapp_regression}. Accordingly, the categories \textit{Not similar}, \textit{Somewhat similar}, \textit{Similar}, and \textit{Very similar} were mapped to $0$, $1/3$, $2/3$, and $1$, respectively. This transformation preserves the ordering of the original ratings while assuming equal numerical spacing between adjacent response categories.

\subsubsection{Feature Generation}

For each color pair, three complementary feature representations were extracted.

The RGB representation was described using the absolute differences between the corresponding color channels:

\begin{equation}
dR=|R_1-R_2|,\quad
dG=|G_1-G_2|,\quad
dB=|B_1-B_2|.
\end{equation}

The colors were additionally converted into the HSI color space, where perceptually meaningful differences were computed. Hue difference was calculated as the minimum circular angular distance,

\begin{equation}
d_H=\frac{\min(|H_1-H_2|,\,360-|H_1-H_2|)}{180},
\end{equation}

while saturation and intensity differences were normalized as

\begin{equation}
d_S=\frac{|S_1-S_2|}{100},\qquad
d_I=\frac{|I_1-I_2|}{255}.
\end{equation}

Finally, each color was represented using the COLIBRI model, consisting of 9 hue categories, 4 saturation levels, and 5 intensity levels. The differences between the corresponding COLIBRI categories were used as additional perceptual features.


Based on these representations, seven feature sets were constructed: RGB, HSI, COLIBRI, RGB+HSI, RGB+COLIBRI, HSI+COLIBRI, and RGB+HSI+COLIBRI.

\subsubsection{Model Training}

The normalized expert similarity scores served as the regression targets. Separate regression models were trained for each feature set to evaluate the predictive capability of the individual color representations and their combinations. This experimental design enables a direct comparison of the contribution of RGB, HSI, and COLIBRI features to the prediction of perceived color similarity.

Model performance was evaluated using five-fold grouped cross-validation, with each unique color pair treated as a grouping variable. All perceptual evaluations corresponding to the same color pair were assigned exclusively to either the training or testing subset within each fold, thereby preventing the same color pair from appearing in both subsets and reducing potential information leakage. Across the five folds, each group was used for test once, while the remaining groups were used for model training.


Model performance was assessed using three standard regression metrics: Mean Squared Error (MSE), Root Mean Squared Error (RMSE), and the coefficient of determination ($R^2$):

\begin{equation}
\mathrm{MSE} = \frac{1}{n}\sum_{i=1}^{n}(y_i-\hat{y}_i)^2,
\end{equation}

\begin{equation}
\mathrm{RMSE} = \sqrt{\frac{1}{n}\sum_{i=1}^{n}(y_i-\hat{y}_i)^2},
\end{equation}

\begin{equation}
R^2 = 1 -
\frac{\sum_{i=1}^{n}(y_i-\hat{y}_i)^2}
{\sum_{i=1}^{n}(y_i-\bar{y})^2}.
\end{equation}

Here, $y_i$ denotes the actual perceptual similarity score, $\hat{y}_i$ is the predicted score, $\bar{y}$ is the mean of the actual scores, and $n$ is the number of observations. MSE and RMSE quantify the prediction error, with lower values indicating better predictive performance, while $R^2$ represents the proportion of variance in the observed scores explained by the model. 

Together, these metrics provide a comprehensive comparison of the different feature representations. The performance of the machine learning models across different feature sets, evaluated in terms of RMSE, MSE, and ($R^2$), is presented in Table \ref{tab:metrics}.

The reported RMSE, MSE, and $R^2$ values represent the mean $\pm$ standard deviation across the five validation folds. The original four-level similarity rating was used only to derive the normalized regression target and was not included among the predictive features.

\subsubsection{Comparative evaluation of models}

\begin{table}[!t]
\centering
\caption{Summary of the perceptual color similarity dataset and grouped cross-validation design.}
\label{tab:dataset_summary}
\small
\renewcommand{\arraystretch}{1.15}

\begin{tabular}{lr}
\toprule
\textbf{Dataset characteristic} & \textbf{Value} \\
\midrule

Participants & 7 \\
Unique color pairs & 2 000 \\
Total evaluations & 28 600 \\

\midrule
\multicolumn{2}{c}{\textbf{Human similarity rating distribution}} \\
\midrule

Not similar & 9 501 (33.22\%) \\
Somewhat similar & 7 192 (25.15\%) \\
Similar & 6 762 (23.64\%) \\
Very similar & 5 145 (17.99\%) \\

\midrule
\multicolumn{2}{c}{\textbf{Grouped cross-validation design}} \\
\midrule

Number of folds & 5 \\
Grouping unit & Unique color pair \\
Training color pairs per fold & 1 600 \\
Validation color pairs per fold & 400 \\
Training evaluations per fold & $\approx$ 22 845 \\
Validation evaluations per fold & $\approx$ 5 755 \\

\bottomrule
\end{tabular}
\end{table}


\begin{table*}[!t]
\centering
\caption{Performance comparison of regression models across different feature sets using five-fold grouped cross-validation. Values are shown as mean $\pm$ standard deviation across folds. }
\label{tab:metrics}
\resizebox{\textwidth}{!}{
\begin{tabular}{@{}llccccc@{}}
\toprule
\textbf{Metric} 
& \textbf{Feature Set} 
& \textbf{Linear Regression} 
& \textbf{Decision Tree} 
& \textbf{Random Forest} 
& \textbf{LightGBM} 
& \textbf{XGBoost} \\

\midrule

\multirow{7}{*}{\textbf{RMSE}}
& RGB
& $0.2653 \pm 0.003$
& $0.2794 \pm 0.006$
& $0.2519 \pm 0.004$
& \textbf{$0.2438 \pm 0.004$}
& $0.2530 \pm 0.005$ \\

& HSI
& $0.2616 \pm 0.004$
& $0.2546 \pm 0.007$
& $0.2312 \pm 0.003$
& \textbf{$0.2268 \pm 0.001$}
& $0.2344 \pm 0.002$ \\

& COLIBRI
& $0.2337 \pm 0.004$
& $0.2654 \pm 0.009$
& $0.2341 \pm 0.006$
& \textbf{$0.2274 \pm 0.005$}
& $0.2364 \pm 0.003$ \\

& RGB + HSI
& $0.2552 \pm 0.004$
& $0.2409 \pm 0.005$
& $0.2155 \pm 0.003$
& \textbf{$0.2116 \pm 0.003$}
& $0.2161 \pm 0.003$ \\

& RGB + COLIBRI
& $0.2248 \pm 0.005$
& $0.2422 \pm 0.003$
& $0.2118 \pm 0.001$
& \textbf{$0.2033 \pm 0.002$}
& $0.2091 \pm 0.002$ \\

& HSI + COLIBRI
& $0.2315 \pm 0.004$
& $0.2413 \pm 0.002$
& $0.2143 \pm 0.002$
& \textbf{$0.2104 \pm 0.002$}
& $0.2156 \pm 0.001$ \\

& RGB + HSI + COLIBRI
& $0.2242 \pm 0.004$
& $0.2305 \pm 0.005$
& $0.2052 \pm 0.002$
& \textbf{$0.2002 \pm 0.002$}
& $0.2036 \pm 0.002$ \\

\midrule

\multirow{7}{*}{\textbf{MSE}}
& RGB
& $0.0704 \pm 0.002$
& $0.0781 \pm 0.003$
& $0.0635 \pm 0.002$
& \textbf{$0.0595 \pm 0.002$}
& $0.0640 \pm 0.003$ \\

& HSI
& $0.0685 \pm 0.002$
& $0.0649 \pm 0.004$
& $0.0535 \pm 0.001$
& \textbf{$0.0514 \pm 0.001$}
& $0.0549 \pm 0.001$ \\

& COLIBRI
& $0.0546 \pm 0.002$
& $0.0705 \pm 0.005$
& $0.0548 \pm 0.003$
& \textbf{$0.0517 \pm 0.002$}
& $0.0559 \pm 0.002$ \\

& RGB + HSI
& $0.0652 \pm 0.002$
& $0.0580 \pm 0.002$
& $0.0465 \pm 0.001$
& \textbf{$0.0448 \pm 0.001$}
& $0.0467 \pm 0.001$ \\

& RGB + COLIBRI
& $0.0506 \pm 0.002$
& $0.0587 \pm 0.002$
& $0.0449 \pm 0.001$
& \textbf{$0.0413 \pm 0.001$}
& $0.0437 \pm 0.001$ \\

& HSI + COLIBRI
& $0.0536 \pm 0.002$
& $0.0582 \pm 0.001$
& $0.0459 \pm 0.001$
& \textbf{$0.0443 \pm 0.001$}
& $0.0465 \pm 0.001$ \\

& RGB + HSI + COLIBRI
& $0.0503 \pm 0.002$
& $0.0532 \pm 0.003$
& $0.0421 \pm 0.001$
& \textbf{$0.0401 \pm 0.001$}
& $0.0415 \pm 0.001$ \\

\midrule

\multirow{7}{*}{\textbf{$R^2$}}
& RGB
& $0.4790 \pm 0.016$
& $0.4212 \pm 0.041$
& $0.5299 \pm 0.023$
& \textbf{$0.5600 \pm 0.015$}
& $0.5260 \pm 0.026$ \\

& HSI
& $0.4932 \pm 0.017$
& $0.5198 \pm 0.026$
& $0.6041 \pm 0.015$
& \textbf{$0.6192 \pm 0.012$}
& $0.5931 \pm 0.014$ \\

& COLIBRI
& $0.5954 \pm 0.023$
& $0.4788 \pm 0.025$
& $0.5946 \pm 0.012$
& \textbf{$0.6171 \pm 0.013$}
& $0.5863 \pm 0.010$ \\

& RGB + HSI
& $0.5177 \pm 0.019$
& $0.5703 \pm 0.022$
& $0.6560 \pm 0.015$
& \textbf{$0.6685 \pm 0.016$}
& $0.6542 \pm 0.015$ \\

& RGB + COLIBRI
& $0.6253 \pm 0.025$
& $0.5658 \pm 0.014$
& $0.6676 \pm 0.014$
& \textbf{$0.6939 \pm 0.014$}
& $0.6763 \pm 0.010$ \\

& HSI + COLIBRI
& $0.6029 \pm 0.025$
& $0.5690 \pm 0.010$
& $0.6598 \pm 0.015$
& \textbf{$0.6719 \pm 0.016$}
& $0.6559 \pm 0.012$ \\

& RGB + HSI + COLIBRI
& $0.6273 \pm 0.023$
& $0.6064 \pm 0.019$
& $0.6882 \pm 0.014$
& \textbf{$0.7031 \pm 0.013$}
& $0.6929 \pm 0.013$ \\
\bottomrule
\end{tabular}
}
\end{table*}

Table~\ref{tab:metrics} compares the predictive performance of the five regression algorithms across the seven feature representations using five-fold grouped cross-validation. The reported values are expressed as mean $\pm$ standard deviation across the validation folds, providing an assessment of both predictive accuracy and stability for unseen color pairs.

A notable result was observed for the COLIBRI representation under Linear Regression. When used as the sole input representation, COLIBRI achieved an average $R^2$ of $0.5954 \pm 0.023$, compared with $0.4790 \pm 0.016$ for RGB and $0.4932 \pm 0.017$ for HSI. This indicates that the fuzzy semantic representation captures a substantially more informative relationship with human similarity judgments under a linear modeling assumption. Combining COLIBRI with RGB further increased the average $R^2$ to $0.6253 \pm 0.025$, while the complete RGB+HSI+COLIBRI representation achieved $R^2=0.6273 \pm 0.023$.

Among the nonlinear algorithms, LightGBM demonstrated the strongest overall predictive performance. Using the complete RGB+HSI+COLIBRI representation, LightGBM achieved the numerically highest average coefficient of determination, $R^2=0.7031 \pm 0.013$, together with RMSE$=0.2002 \pm 0.002$ and MSE$=0.0401 \pm 0.001$. The low variability across folds indicates stable performance across different subsets of previously unseen color pairs. LightGBM also benefited consistently from combining multiple feature representations, with $R^2$ increasing from $0.5600$ for RGB alone to $0.6939$ for RGB+COLIBRI and $0.7031$ for the complete feature set.



Table~\ref{tab:feature_importance} presents the Linear Regression parameters for each feature representation, including the intercept and slope coefficients. Each coefficient quantifies the direction and magnitude of the contribution of the corresponding color-difference feature.

\begin{table*}[t]
\caption{Linear regression coefficients for perceptual color similarity across different feature representations using five-fold grouped cross-validation. Values are shown as mean $\pm$ standard deviation across folds.}
\label{tab:feature_importance}
\centering

\resizebox{\textwidth}{!}{%
\begin{tabular}{@{}llccccccc@{}}
\toprule
\textbf{Model Dependency} 
& \textbf{Feature} 
& \multicolumn{7}{c}{\textbf{Coefficient (Mean $\pm$ SD)}} \\
\cmidrule(lr){3-9}

&
&
\textbf{RGB}
& \textbf{HSI}
& \textbf{COLIBRI}
& \textbf{RGB + HSI}
& \textbf{RGB + COLIBRI}
& \textbf{HSI + COLIBRI}
& \textbf{RGB + HSI + COLIBRI} \\
\midrule

{}
& \textbf{Intercept}
& $0.6930 \pm 0.005$
& $0.6792 \pm 0.005$
& $0.7851 \pm 0.005$
& $0.6954 \pm 0.005$
& $0.7899 \pm 0.005$
& $0.7799 \pm 0.005$
& $0.7937 \pm 0.005$ \\

\midrule

{}
& \textbf{Red Channel}
& $-0.0027 \pm 0.0$
& -
& -
& $-0.0014 \pm 0.0$
& $-0.0015 \pm 0.0$
& -
& $-0.0018 \pm 0.0$ \\

\textbf{RGB}
& \textbf{Green Channel}
& $-0.0029 \pm 0.0$
& -
& -
& $-0.0016 \pm 0.0$
& $-0.0013 \pm 0.0$
& -
& $-0.0016 \pm 0.0$ \\

{}
& \textbf{Blue Channel}
& $-0.0024 \pm 0.0$
& -
& -
& $-0.0013 \pm 0.0$
& $-0.0011 \pm 0.0$
& -
& $-0.0014 \pm 0.000$ \\

\midrule

{}
& \textbf{Hue}
& -
& $-0.7848 \pm 0.007$
& -
& $-0.4954 \pm 0.010$
& -
& $-0.2282 \pm 0.008$
& $0.0950 \pm 0.016$ \\

\textbf{HSI}
& \textbf{Saturation}
& -
& $-0.2406 \pm 0.011$
& -
& $-0.0902 \pm 0.011$
& -
& $-0.1117 \pm 0.025$
& $0.1104 \pm 0.021$ \\

{}
& \textbf{Intensity}
& -
& $-0.8270 \pm 0.038$
& -
& $-0.1712 \pm 0.041$
& -
& $-0.2042 \pm 0.078$
& $0.6119 \pm 0.114$ \\

\midrule

{}
& \textbf{Hue -- Blue}
& -
& -
& $-0.2608 \pm 0.010$
& -
& $-0.1951 \pm 0.007$
& $-0.2087 \pm 0.009$
& $-0.2026 \pm 0.008$ \\

{}
& \textbf{Hue -- Cyan}
& -
& -
& $-0.2329 \pm 0.011$
& -
& $-0.1666 \pm 0.005$
& $-0.1956 \pm 0.009$
& $-0.1700 \pm 0.005$ \\

{}
& \textbf{Hue -- Green}
& -
& -
& $-0.3199 \pm 0.003$
& -
& $-0.2121 \pm 0.003$
& $-0.2379 \pm 0.004$
& $-0.2234 \pm 0.004$ \\

{}
& \textbf{Hue -- Light Blue}
& -
& -
& $-0.2005 \pm 0.005$
& -
& $-0.2004 \pm 0.005$
& $-0.1928 \pm 0.005$
& $-0.2013 \pm 0.005$ \\

{}
& \textbf{Hue -- Magenta}
& -
& -
& $-0.2685 \pm 0.004$
& -
& $-0.2062 \pm 0.002$
& $-0.2126 \pm 0.004$
& $-0.2173 \pm 0.004$ \\

{}
& \textbf{Hue -- Orange}
& -
& -
& $-0.2503 \pm 0.004$
& -
& $-0.2111 \pm 0.005$
& $-0.2247 \pm 0.004$
& $-0.2147 \pm 0.006$ \\

{}
& \textbf{Hue -- Red}
& -
& -
& $-0.2436 \pm 0.011$
& -
& $-0.2095 \pm 0.009$
& $-0.2186 \pm 0.009$
& $-0.2111 \pm 0.009$ \\

{}
& \textbf{Hue -- Violet}
& -
& -
& $-0.2746 \pm 0.007$
& -
& $-0.2072 \pm 0.004$
& $-0.2216 \pm 0.006$
& $-0.2131 \pm 0.007$ \\

{}
& \textbf{Hue -- Yellow}
& -
& -
& $-0.3266 \pm 0.018$
& -
& $-0.2874 \pm 0.015$
& $-0.2989 \pm 0.015$
& $-0.2964 \pm 0.015$ \\

\textbf{COLIBRI}
& \textbf{Saturation -- High}
& -
& -
& $-0.0219 \pm 0.004$
& -
& $0.0497 \pm 0.006$
& $0.0320 \pm 0.009$
& $0.0208 \pm 0.007$ \\

{}
& \textbf{Saturation -- Low}
& -
& -
& $-0.1219 \pm 0.007$
& -
& $-0.1029 \pm 0.004$
& $-0.0827 \pm 0.009$
& $-0.1322 \pm 0.007$ \\

{}
& \textbf{Saturation -- Medium}
& -
& -
& $-0.0762 \pm 0.006$
& -
& $-0.0836 \pm 0.005$
& $-0.0880 \pm 0.007$
& $-0.0802 \pm 0.005$ \\

{}
& \textbf{Saturation -- Very Low}
& -
& -
& $-0.0705 \pm 0.013$
& -
& $-0.1146 \pm 0.020$
& $-0.0118 \pm 0.018$
& $-0.1786 \pm 0.022$ \\

{}
& \textbf{Intensity -- Black}
& -
& -
& $-0.0791 \pm 0.013$
& -
& $-0.0143 \pm 0.014$
& $-0.0125 \pm 0.025$
& $-0.1335 \pm 0.026$ \\

{}
& \textbf{Intensity -- Dark gray}
& -
& -
& $-0.0789 \pm 0.014$
& -
& $-0.0027 \pm 0.012$
& $-0.0499 \pm 0.016$
& $-0.0610 \pm 0.017$ \\

{}
& \textbf{Intensity -- Gray}
& -
& -
& $-0.1338 \pm 0.011$
& -
& $-0.0678 \pm 0.011$
& $-0.1040 \pm 0.014$
& $-0.0964 \pm 0.012$ \\

{}
& \textbf{Intensity -- Light gray}
& -
& -
& $-0.1798 \pm 0.014$
& -
& $0.0154 \pm 0.013$
& $-0.1133 \pm 0.016$
& $-0.1268 \pm 0.027$ \\

{}
& \textbf{Intensity -- White}
& -
& -
& $-0.0925 \pm 0.022$
& -
& $-0.1117 \pm 0.057$
& $-0.0757 \pm 0.038$
& $-0.1330 \pm 0.038$ \\

\bottomrule
\end{tabular}%
}

\end{table*}

The predicted perceptual similarity is expressed as
\begin{equation}
\hat{S}_{\mathrm{perc}} =
\beta_0 + \sum_{j=1}^{24} \beta_j x_j,
\label{eq:perceptual_similarity}
\end{equation}
where $\beta_0$ denotes the intercept, $\beta_j$ is the regression coefficient associated with the $j$-th feature, and $x_j$ represents the corresponding color-difference feature.

The  perceptual color difference is calculated as
\begin{equation}
\hat{D}_{\mathrm{perc}} =
1-\hat{S}_{\mathrm{perc}},
\label{eq:perceptual_difference}
\end{equation}


\subsubsection{Experimental results analysis}

- Analysis of experimental findings.
-Outcomes

\begin{figure}
    \centering
    \includegraphics[width=0.9\linewidth]{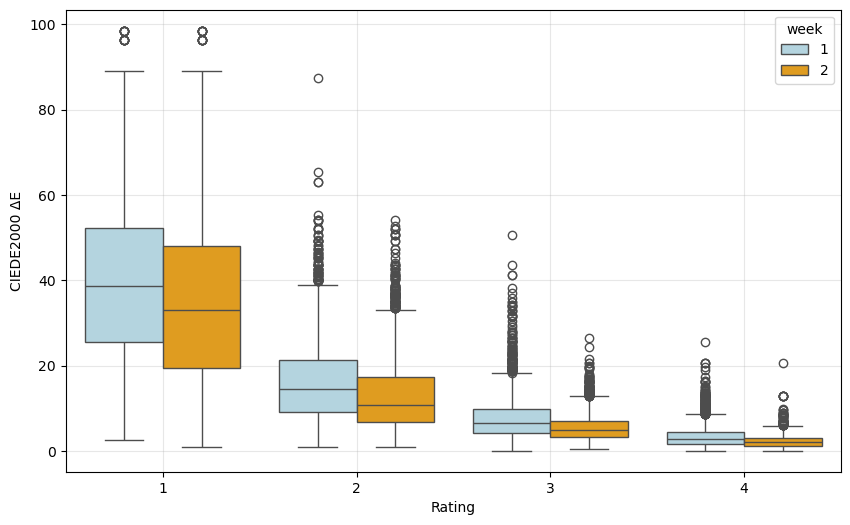}
    \caption{Experimental results for Experiment 1 and 2 side to side}
    \label{fig:exp_results_boxplot}
\end{figure}

Figure \ref{fig:exp_results_boxplot} shows us results of both experiments as boxplots as side by side. Clearly can be seen that regardless of the experiments $\Delta E$ of "Not similar" are similar by results of the both experiments. That result brings us to the hypothesis that it is easier for people to say whenever two colors are different than similar. 

The correlation between participants' ratings across the two experiments was 0.817. To further evaluate reliability of similarity judgments, we computed the Intraclass Correlation Coefficient (ICC). The obtained value ICC(2,1) = 0.826 (95\% CI [0.77, 0.87], p $<$ 0.001) indicates good agreement between participants ( reference ICC paper). Ratings of duplicated color pairs in the second experiment were excluded from the calculation.

\begin{table}[h]
\caption{Perceptual similarity decision boundaries thresholds}
\begin{tabular}{lcc}
\toprule
\textbf{Category} & \textbf{Experiment 1} & \textbf{Experiment 2 } \\
\midrule
Very Similar & $\Delta E \leq 4.221$ & $\Delta E \leq 2.069$ \\
Similar & $4.221 < \Delta E \leq 11.406$ & $2.069 < \Delta E \leq 6.743$ \\
Somewhat Similar & $11.406 < \Delta E \leq 24.266$ & $6.743 < \Delta E \leq 18.889$ \\
Not Similar & $\Delta E > 24.266$ & $\Delta E > 18.889$ \\
\bottomrule
\end{tabular}
\label{decision_boundaries_experiment}
\end{table}

To obtain numerical boundaries between perceptual similarity categories, we fitted logistic regression models using the color difference $\Delta E$ as the predictor and binarized rating thresholds (rating $\geq$ 4, $\geq$ 3, $\geq$ 2) as target variables, results are illustrated in Table \ref{decision_boundaries_experiment}.

To compare how participants voted to the experiments we calculated Fleiss' Kappa agreement score for both experiments. Following agreement scores were obtained: 0.679 and 0.417 for the first and second experiments respectfully.

\subsection{Performance Evaluation}

\begin{figure*}[!t]
    \centering
    \includegraphics[width=0.85\linewidth]{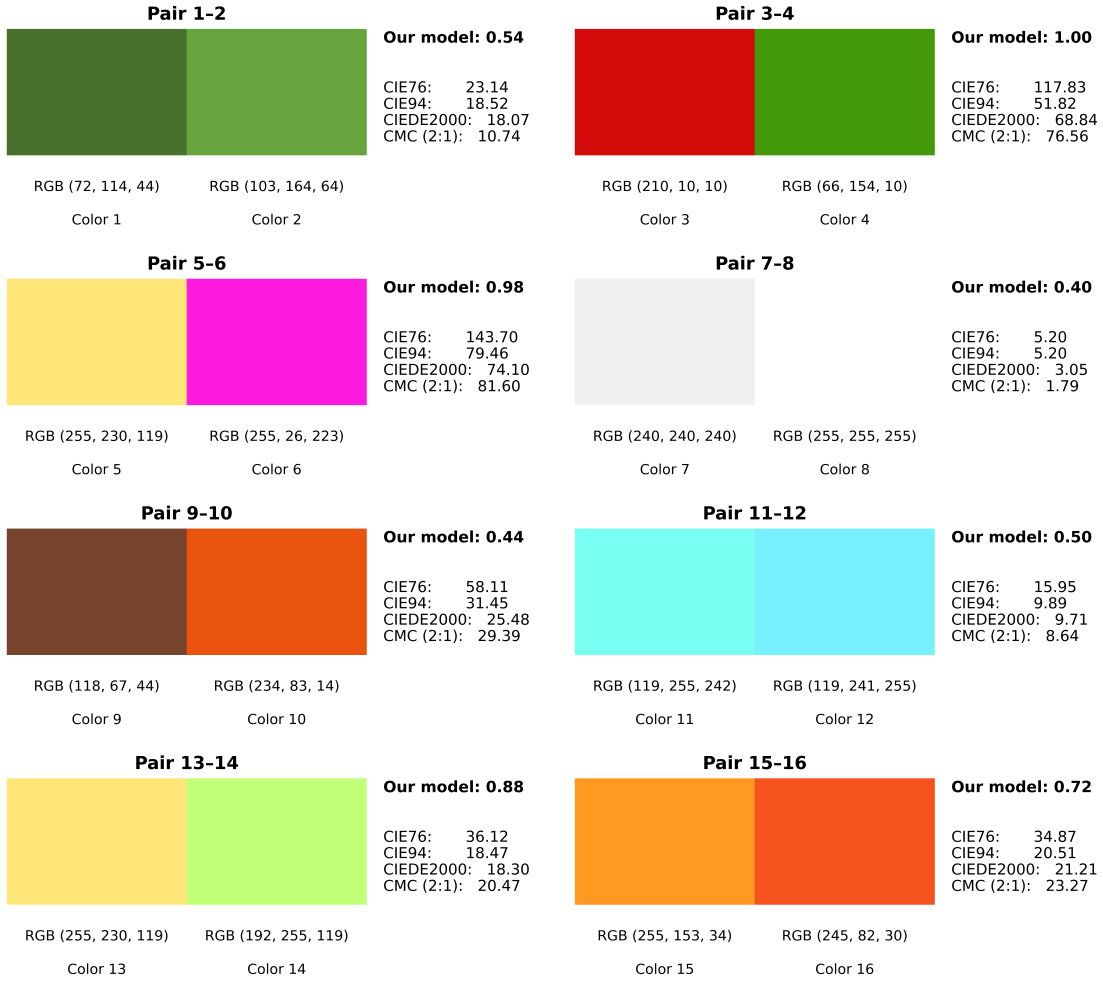}
    \caption{Comparison of the proposed model with conventional color-difference measures.}
    \label{fig:color_difference_comparison}
\end{figure*}

To further examine the perceptual behavior of the proposed approach, we compared its predictions with conventional color-difference measures, including CIE76, CIE94, CIEDE2000, and CMC ($2{:}1$). Representative color pairs and the corresponding metric values are illustrated in Figure~\ref{fig:color_difference_comparison}.

The proposed model produces a normalized color-difference score
$D_{\mathrm{model}}\in[0,1]$, where larger values indicate a greater perceived difference. 
For interpretation in terms of perceptual similarity, the predicted difference was converted as

\begin{equation}
S_{\mathrm{model}} = 1 - D_{\mathrm{model}}.
\end{equation}

The resulting similarity score was then mapped back to the four linguistic categories used in the perceptual experiment. According to the adopted normalization, \textit{Not similar}, \textit{Somewhat similar}, \textit{Similar}, and \textit{Very similar} correspond to the anchor values $0$, $1/3$, $2/3$, and $1$, respectively. Each continuous prediction was assigned to the nearest anchor value. Therefore, the category boundaries were defined at the midpoints between adjacent anchors: $1/6\approx0.167$, $1/2=0.500$, and $5/6\approx0.833$. Thus, $S_{\mathrm{model}}<0.167$ was interpreted as \textit{Not similar}, $0.167\leq S_{\mathrm{model}}<0.500$ as \textit{Somewhat similar}, $0.500\leq S_{\mathrm{model}}<0.833$ as \textit{Similar}, and $S_{\mathrm{model}}\geq0.833$ as \textit{Very similar}.

For comparison, CIEDE2000 was interpreted using conventional linguistic color-difference ranges, from \textit{Hardly} perceptible to \textit{Strongly} different. Table~\ref{tab:linguistic_comparison} presents the numerical and linguistic interpretations produced by the proposed our model and CIEDE2000 for the selected color pairs.

\begin{table*}[!t]
\centering
\caption{Linguistic comparison between the proposed model and CIEDE2000.}
\label{tab:linguistic_comparison}
\renewcommand{\arraystretch}{1.15}
\setlength{\tabcolsep}{5pt}

\begin{tabular}{cccccc}
\toprule
\textbf{Pair} &
\textbf{Model Difference} &
\textbf{Model Similarity} &
\textbf{Model Interpretation} &
$\mathbf{\Delta E_{00}}$ &
\textbf{CIEDE2000 Interpretation} \\
\midrule

1--2
& 0.539
& 0.461
& Somewhat similar
& 18.066
& Very much \\

3--4
& 1.000
& 0.000
& Not similar
& 68.840
& Strongly \\

5--6
& 0.982
& 0.018
& Not similar
& 74.104
& Strongly \\

7--8
& 0.403
& 0.597
& Similar
& 3.047
& Appreciable \\

9--10
& 0.444
& 0.556
& Similar
& 25.478
& Strongly \\

11--12
& 0.498
& 0.502
& Similar
& 9.713
& Much \\

13--14
& 0.878
& 0.122
& Not similar
& 18.303
& Very much \\

15--16
& 0.724
& 0.276
& Somewhat similar
& 21.214
& Very much \\

\bottomrule
\end{tabular}
\end{table*}





\section{Conclusion}

This study presents a perceptual color difference through human similarity judgments and machine learning. Two controlled experiments were conducted using 2,000 unique color pairs under separated and no-separation presentation conditions, resulting in 28,600 individual evaluations from seven observers. 

The results show that perceived color similarity depends not only on color distance but also on spatial presentation. In particular, the "Very Similar" boundary decreased from $\Delta E \leq 4.221$ under separation to $\Delta E \leq 2.069$ under no separation, demonstrating a substantial shift in perceptual similarity thresholds. Machine learning experiments further showed that combining complementary color representations improves prediction of human judgments. LightGBM using RGB+HSI+COLIBRI features achieved the best performance ($R^2=0.7031$, RMSE$=0.2002$). Notably, COLIBRI alone outperformed RGB and HSI under linear regression. 



The main limitations are the relatively small observer group, the assumption of equal spacing between ordinal similarity categories, and the use of web-based display conditions. Besides this, a limitation of the regression formulation is that the four-point similarity scale is ordinal by design, whereas its normalization to $[0,1]$ assumes equal numerical spacing between adjacent categories. Therefore, the transformation should be interpreted as a practical modeling approximation rather than evidence that the perceptual distance between consecutive rating categories is inherently equal.

Future work will involve larger-scale perceptual experiments, controlled display conditions, and adaptive neuro-fuzzy approaches for learning perceptual similarity directly from human judgments.










\section*{Acknowledgement}
This research has been funded by the Science Committee of the Ministry of Science and Higher Education of the Republic of Kazakhstan (Grant No. AP22786412)

\begin{IEEEbiography}[{\includegraphics[width=1in,height=1.25in,clip,keepaspectratio]{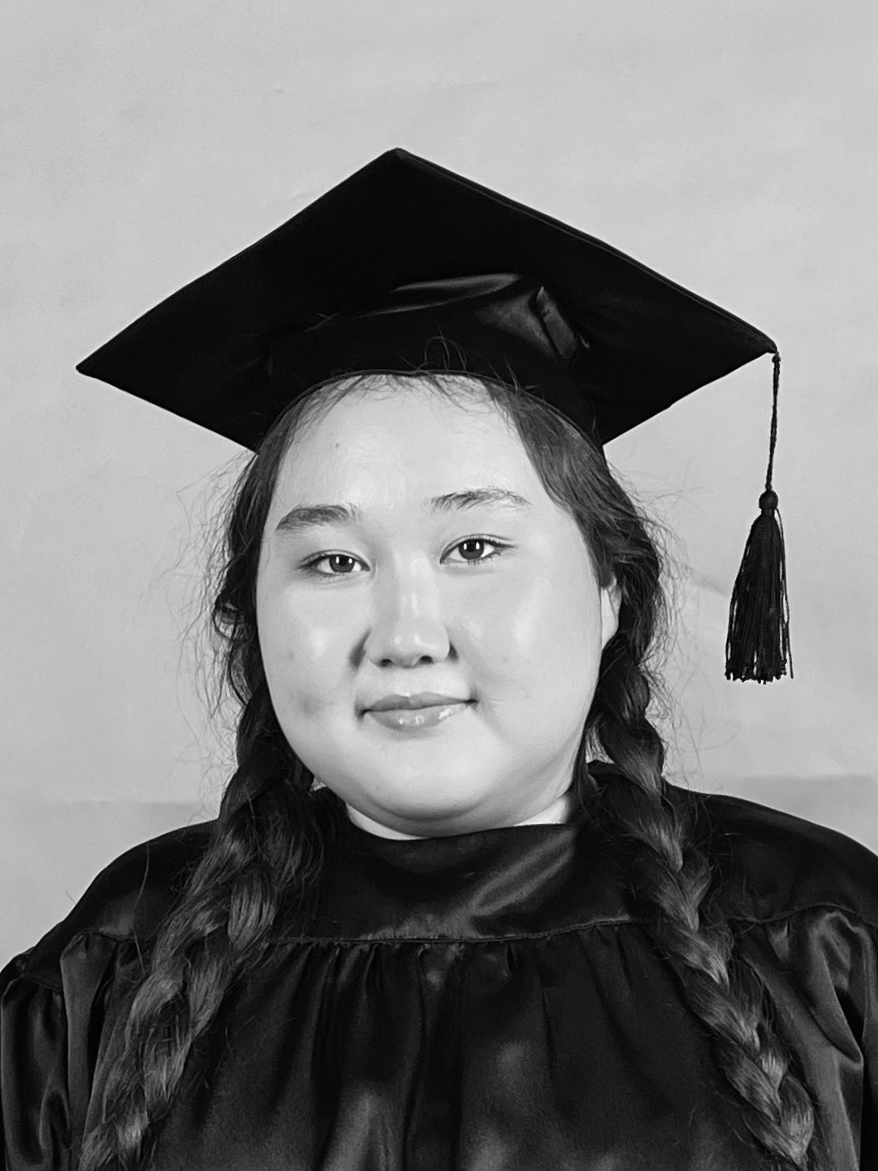}}]{Elnara Kadyrgali} received a B.S. degree in Computer System and Software and M.S. degree in IT management from the Kazakh-British Technical University (KBTU), Almaty, Kazakhstan, in 2022, and 2024 respectively. She is currently pursuing a PhD degree and working as a Senior Researcher at the same university, while also holding a position as a Systems Analyst. She participated in a number of conferences, including SIST 2026, FSDM 2024, IITU YDF-2024, KBTU AGSRW 2023, and 2024 IEEE AITU: Digital Generation, which received the best paper award. She has been serving as a Teaching Assistant at KBTU since Sep. 2024.
Her research interests include colors, emotion recognition and recommendation systems.
\end{IEEEbiography}

\begin{IEEEbiography}[{\includegraphics[width=1in,height=1.25in,clip,keepaspectratio]{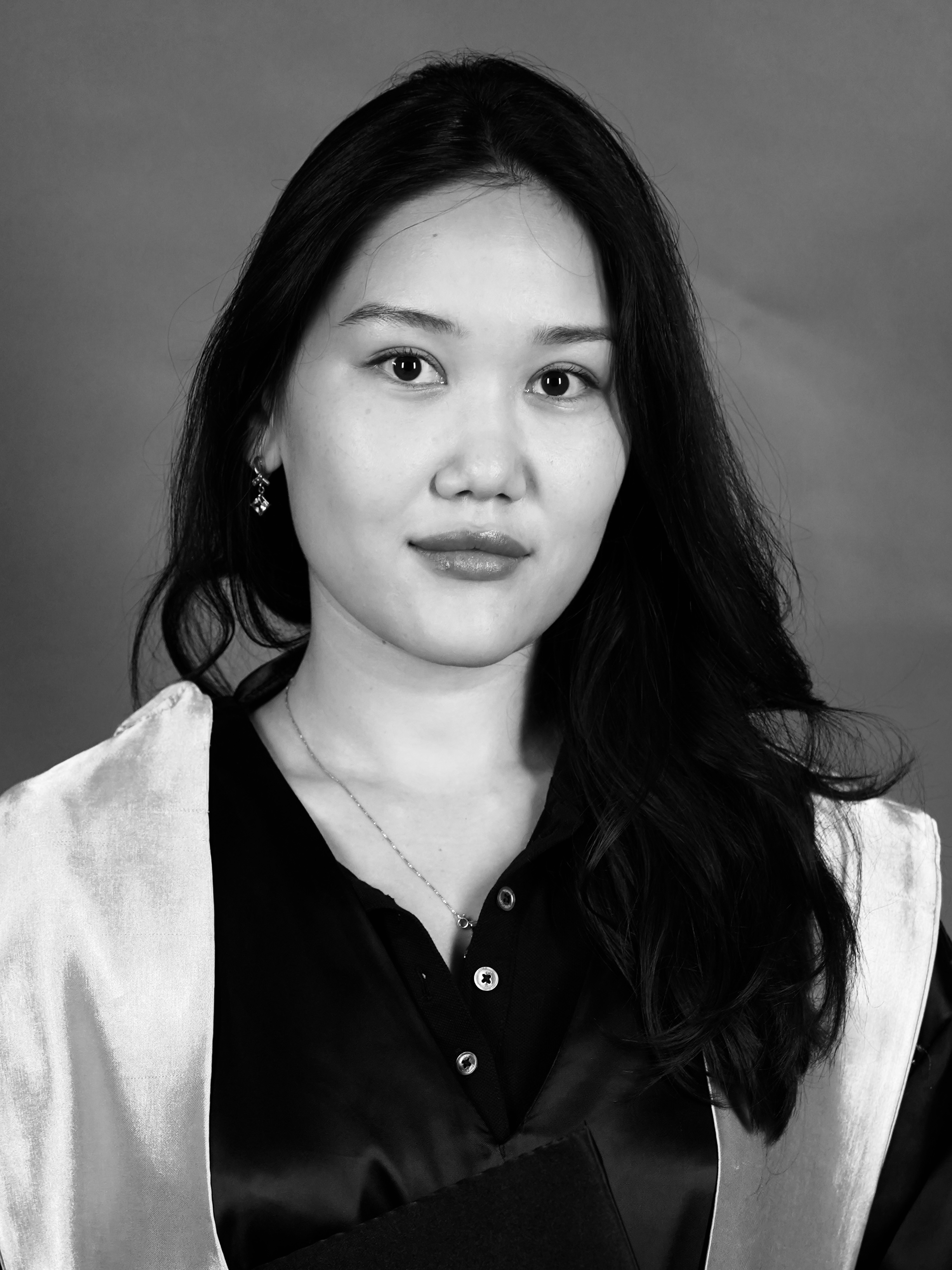}}]{Muragul Muratbekova} received the B.S. degree in Information Systems in 2022 and the M.S. degree in IT Management in 2024, both from the Kazakh-British Technical University (KBTU), Almaty, Kazakhstan. She is currently a Research Associate at KBTU and works as a Senior Software Developer in a leading telecommunication company in Kazakhstan. She has participated in multiple international conferences (EUSPN 2023, FSDM 2023, SCIS\&ISIS 2024), receiving the Best Student Presentation Award at SCIS\&ISIS 2024. She has been serving as a senior lecturer in the KBTU since Sep. 2025. Her research interests include visual perception computing, emotion engineering, and image processing.

\end{IEEEbiography}

\begin{IEEEbiography}[{\includegraphics[width=1in,height=1.25in,clip,keepaspectratio]{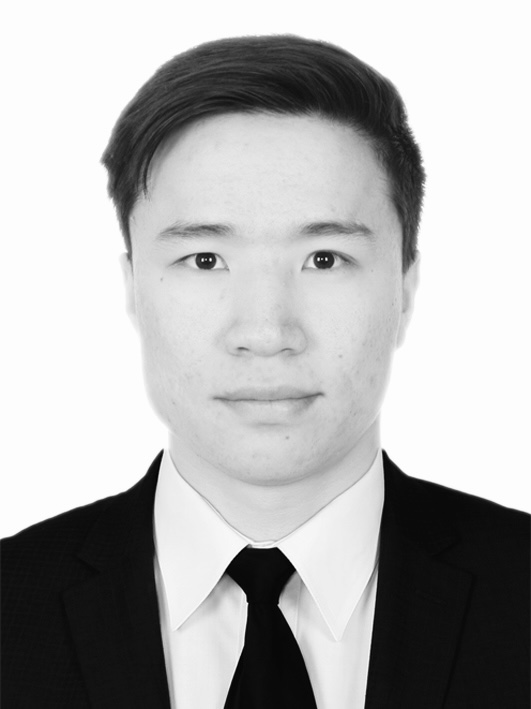}}]{Adilet Yerkin} received a B.S. degree in Information systems from the International University of Information Technologies, Almaty, Kazakhstan, in 2022 and M.S. degree in Data Science with the School of Information Technology and Engineering, at the Kazakh-British Technical University (KBTU), Almaty in 2024. He is currently pursuing a PhD degree at the same university. He participated in conferences, such as FUZZ-IEEE 2025, FSDM 2024, KBTU AGSRW 2023, IEEE AITU 2024 and IITU YDF-2024, which received awards for the best paper. He has been serving as a senior lecturer in the KBTU since Sep. 2025. Also, he works as the Head of the Data Science Division in a state company. His research interests include machine learning, group decision making systems, fuzzy logic and sets.

\end{IEEEbiography}

\begin{IEEEbiography}[{\includegraphics[width=1in,height=1.25in,clip,keepaspectratio]{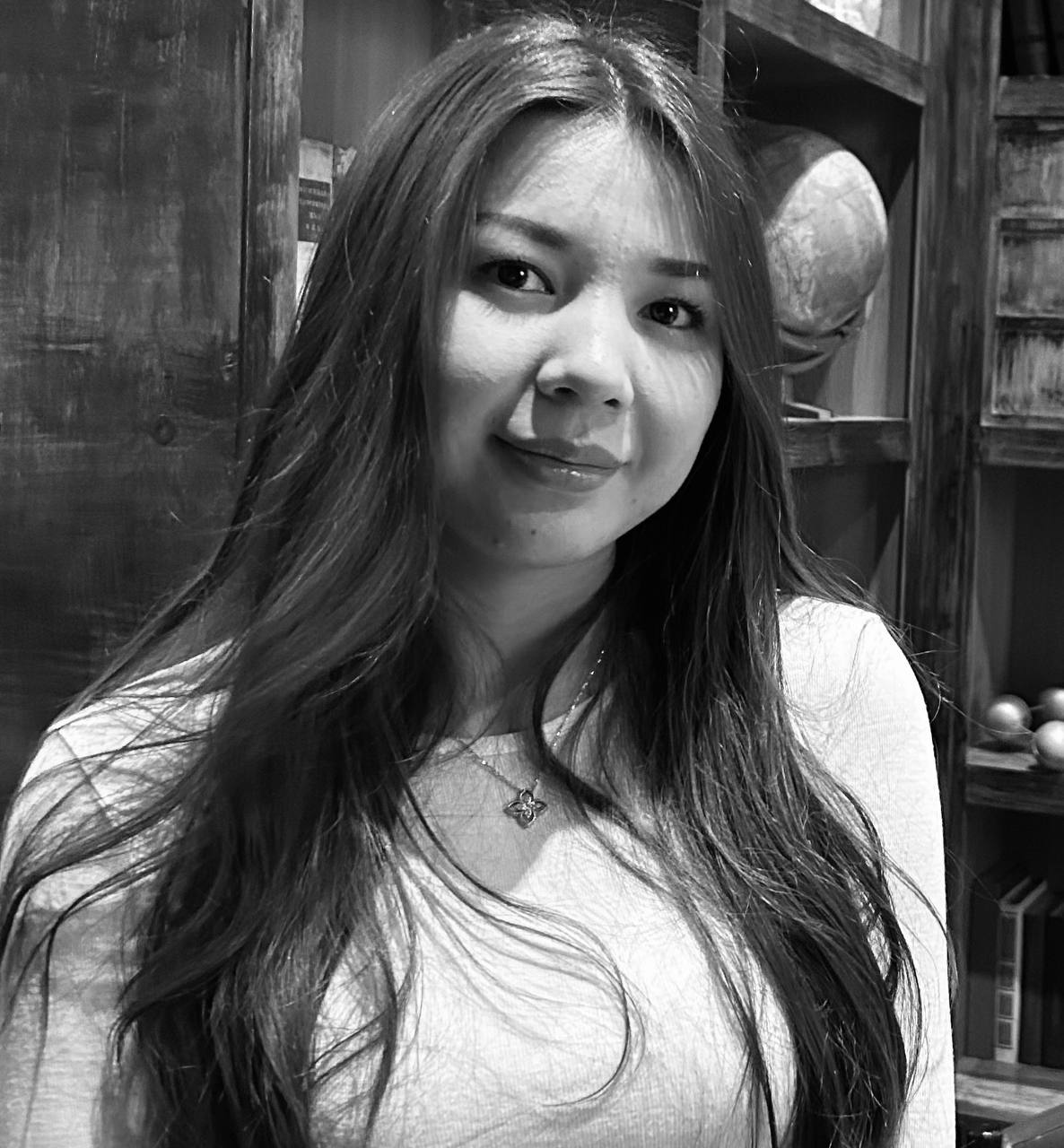}}]{Nuray Toganas} received her B.S. degree in Mathematical and Computer Modeling from Al-Farabi Kazakh National University, Almaty, Kazakhstan, in 2021, and her M.S. degree in IT Management from Kazakh-British Technical University (KBTU), Almaty, Kazakhstan, in 2025. She has been a researcher at KBTU since 2024. She is currently Deputy Dean of the School of IT and Engineering at KBTU. Her research interests include color perception and categorization, fuzzy color modeling, color naming, and machine learning applications in emotion recognition and digital art.
\end{IEEEbiography}

\begin{IEEEbiography}[{\includegraphics[width=1in,height=1.25in,clip,keepaspectratio]{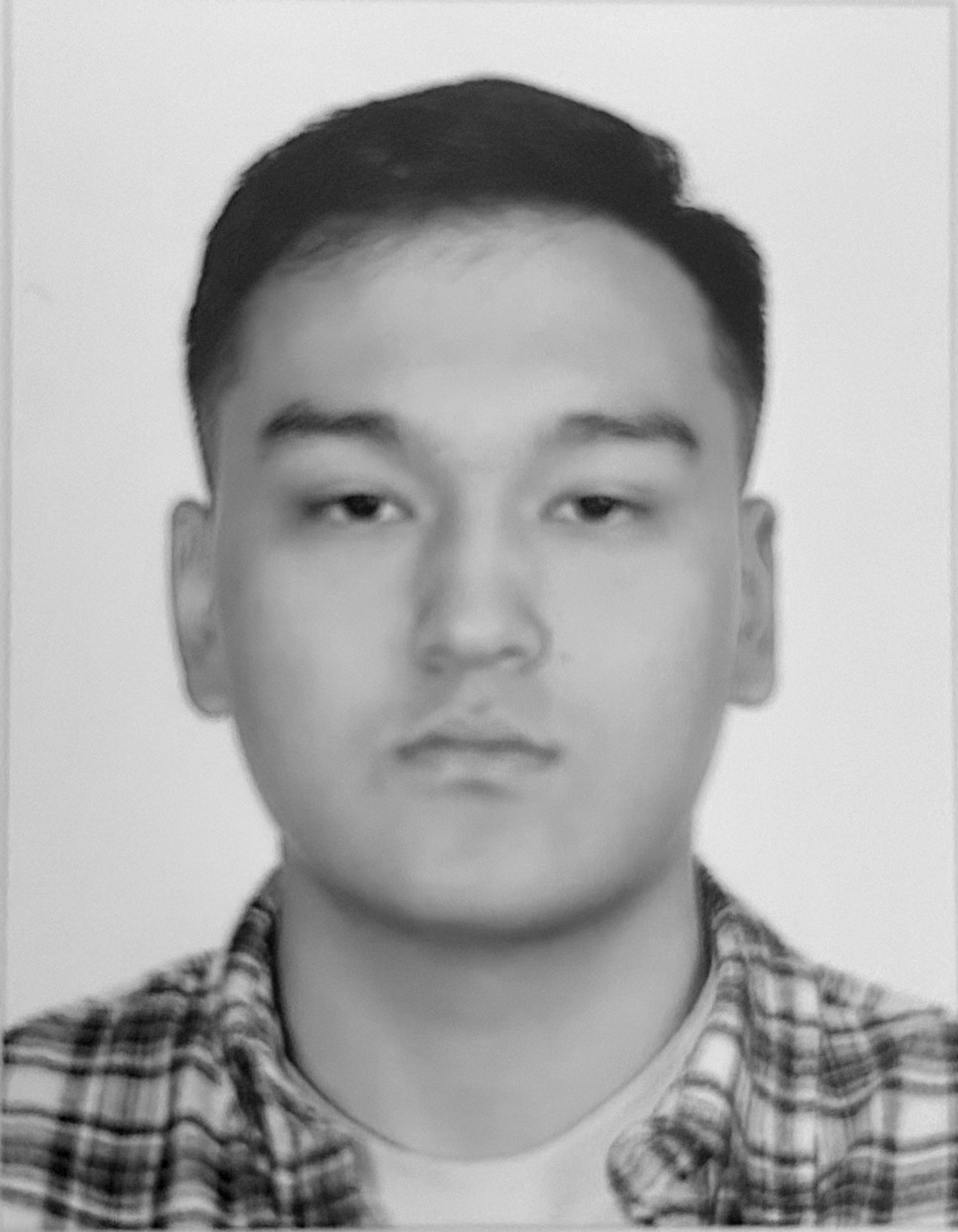}}]{Ayan Igali} received a B.S degree in Information Systems and currently pursing a M.S. degree in Data Science at Kazakh-British Technical University, Almaty, Kazakhstan. He is currently a Researcher at KBTU and works as Middle Machine Learning Engineer in the one of the biggest banks in Kazakhstan. He participated in international conference SCIS\&ISIS 2024. His research interests include machine learning, NLP, computer vision, image processing, color research, emotion detection and human-friendly systems.
\end{IEEEbiography}

\begin{IEEEbiography}[{\includegraphics[width=1in,height=1.25in,clip,keepaspectratio]{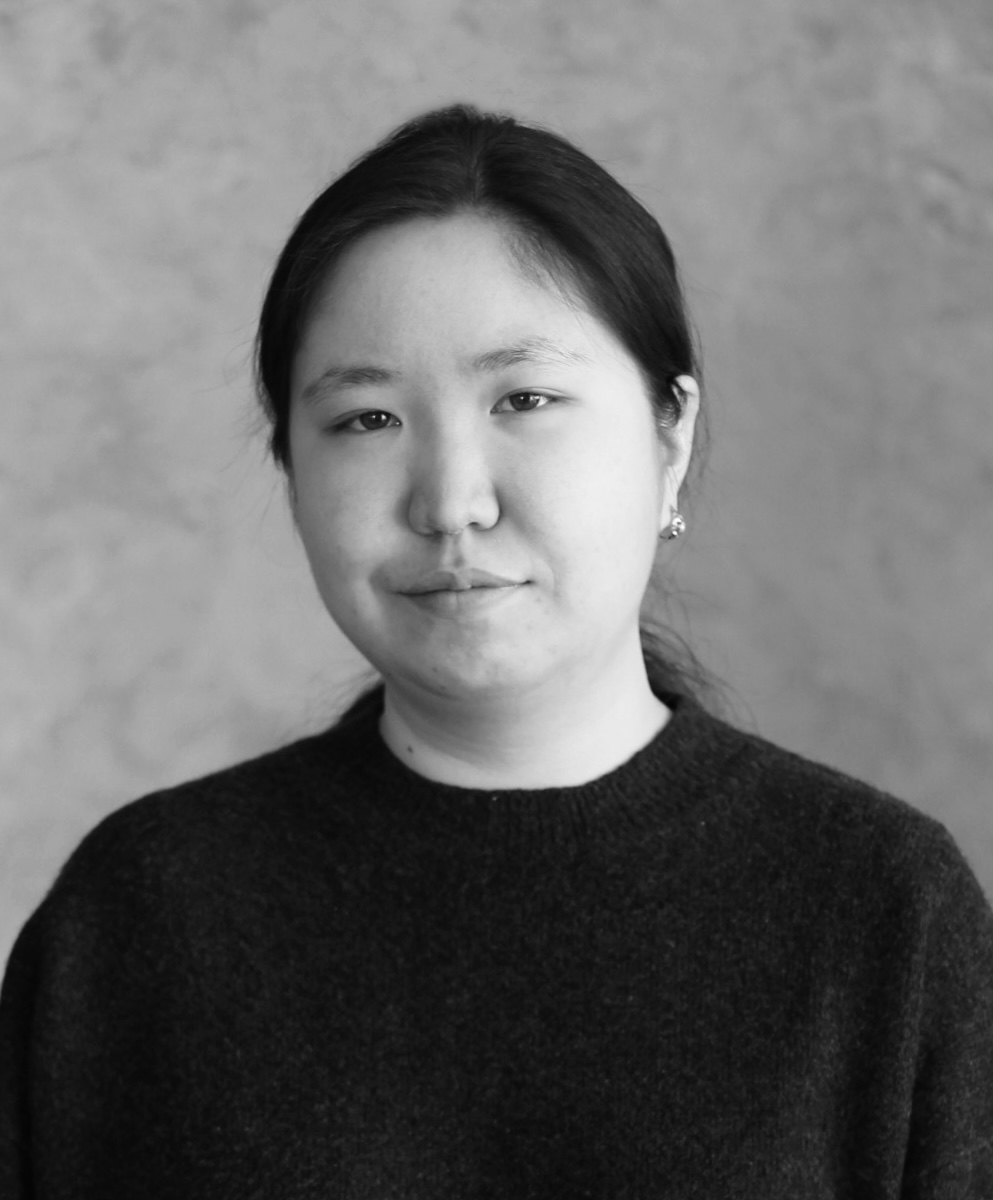}}]{Malika Ziyada} received a B.S. degree in Computer Science from Nazarbayev University, Astana Kazakhstan, in 2023. She has also received an M.S. degree in Data Science at Kazakh-British Technical University in Almaty while working as a junior researcher. She also works as an analyst in a national airline. Her research interests include emotion detection, NLP, machine learning, and visual perception. 
\end{IEEEbiography}

\begin{IEEEbiography}[{\includegraphics[width=1in,height=1.25in,clip,keepaspectratio]{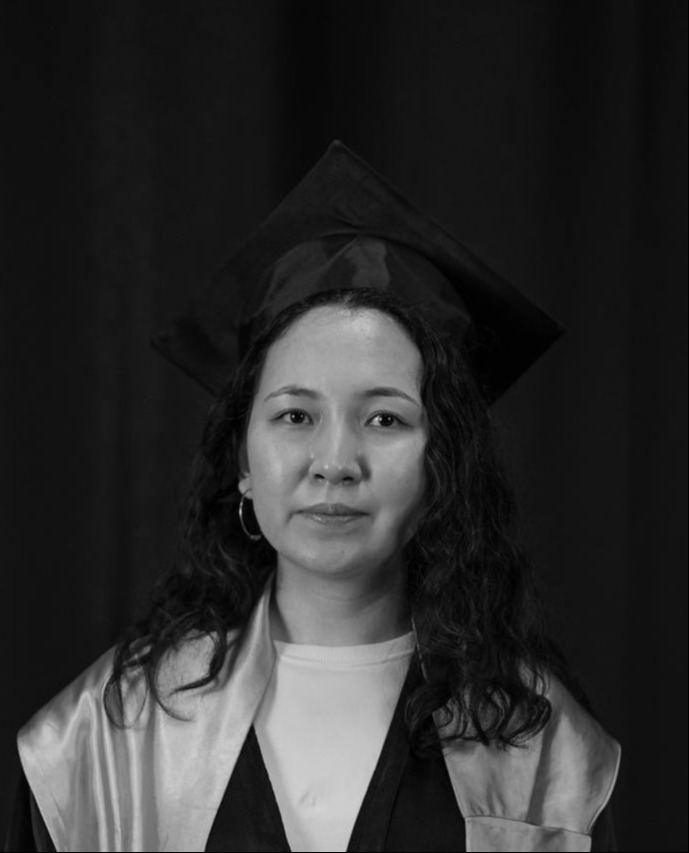}}]{Aruzhan Burambekova} received a B.S. degree in Computational Mathematics and Cybernetics from Kazakhstan Branch of Lomonosov Moscow State University, Astana Kazakhstan, in 2023. She has also received an M.S. degree in Data Science at Kazakh-British Technical University in Almaty, Kazakhstan. She also works as a product analyst at one of the largest e-commerce companies in Kazakhstan. Her research interests include color models, computer vision, machine learning and NLP. 
\end{IEEEbiography}

\begin{IEEEbiography}[{\includegraphics[width=1in,height=1.25in,clip,keepaspectratio]{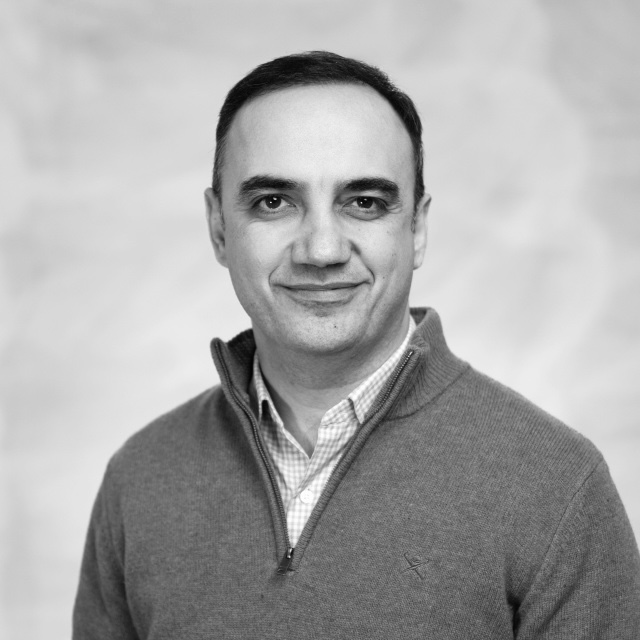}}]{Jamaladdin Hasanov} is currently an Associate Professor of Computer and Information Sciences with the School of IT and Engineering, ADA
University, Baku, Azerbaijan. His primary teaching covers computer-vision and industry-oriented computer science courses. He is also a Fellow
Researcher with MegaSec AI Laboratory, where he leads computer vision-related project, including face identification and recognition, outdoor scene text recognition, and model optimization. His research interests include color vision, computer vision, video captioning, and medical image classification.
\end{IEEEbiography}

\begin{IEEEbiography}[{\includegraphics[width=1in,height=1.25in,clip,keepaspectratio]{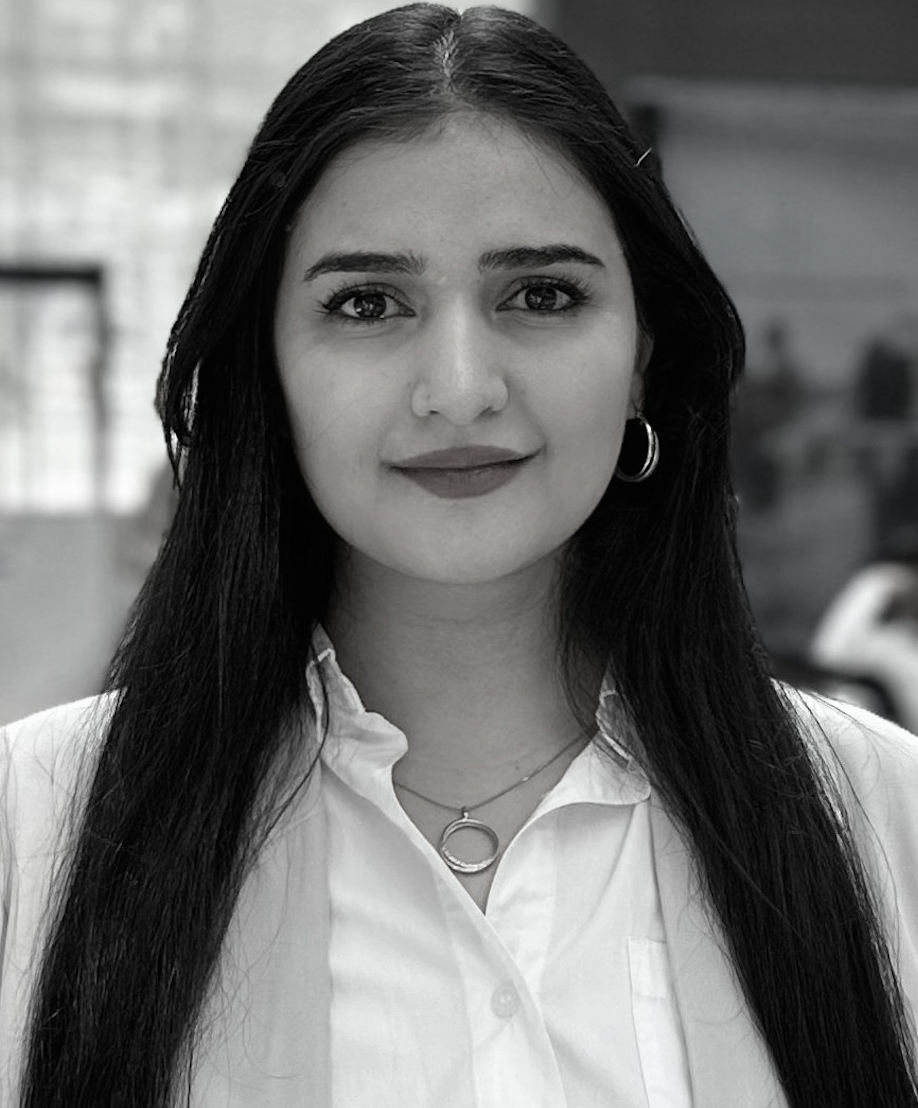}}]{Pakizar Shamoi}  received the B.S. and M.S. degrees in information systems from the Kazakh-British Technical University (KBTU), Almaty, Kazakhstan, in 2011 and 2013, and the Ph.D. degree in engineering from Mie University, Tsu, Japan, in 2019. In her academic journey, she has held various teaching and research positions at KBTU, where she has been serving as a professor in the School of Information Technology and Engineering since August 2020. She is also the head of the FLIS (Fuzzy Logic and Intelligent Systems) Lab and the Computer Science division at the school. She is the author of one book, one monograph, and more than 50 publications. Awards for the best paper at conferences were received five times. Her research interests include AI, focusing on fuzzy sets and logic, soft computing, and the representation and processing of colors and emotions in computer systems. She took part and worked in the org. committee of several international conferences - IFSA-SCIS 2017, Otsu, Japan; SCIS-ISIS 2022, Mie, Japan; EUSPN 2023. She serves as a reviewer at several international conferences and journals.

\end{IEEEbiography}

\bibliographystyle{IEEEtran}
\bibliography{access}

\end{document}